\documentclass[10pt]{article}

\usepackage[pdftex]{graphicx}
\usepackage{amsmath}
\usepackage{amssymb}
\usepackage{color}
\usepackage{algorithm}
\usepackage{multicol}
\usepackage{multirow}
\usepackage{multicol}
\usepackage{array}
\usepackage{fullpage}
\usepackage{algorithmic}

\usepackage{xpatch}

\makeatletter
\xpatchcmd{\algorithmic}{\itemsep\z@}{\itemsep=3pt}{}{}
\makeatother

\newenvironment{et}{\color[rgb]{0,0.5,0}}{}{}
\newcommand{\bet}{\begin{et}}
\newcommand{\eet}{\end{et}}
\newcommand\blfootnote[1]{%
  \begingroup
  \renewcommand\thefootnote{}\footnote{#1}%
  \addtocounter{footnote}{-1}%
  \endgroup
}

 \def\x{{\mathbf x}}

\def\z{{\mathbf z}}
\def\y{{\mathbf y}}
\def\yji{{\mathbf y}_{ji}}
\def\by{\overline{\mathbf y}}
\def\byji{\overline{\mathbf y}_{ji}}
\def\zji{z_{ji}}
\def\alphajik{\alpha_{jik}}
\def\W{{\mathbf W}}
\def\R{{\mathbf R}}
\def\I{{\mathbf I}}

\def\R{{\mathbf R}}
\def\t{{\mathbf t}}

\def\mN{{\mathcal N}}

\def\vji{{\mathbf v}_{ji}}

\def\bmu{\boldsymbol{\mu}}
\def\bSigma{\boldsymbol{\Sigma}}

\def\bSigmaji{\overline{\boldsymbol{\Sigma}}_{ji}}

\def\trace{\text{tr}}

\newcommand\argmax[1]{\underset{#1}{\text{argmax}}\,}
\newcommand\argmin[1]{\underset{#1}{\text{argmin}}\,}

\begin{document}
\title{Multiview point cloud registration with  anisotropic and space-varying localization noise
\blfootnote{This work was supported by the French National Research Agency (ANR) through the SP-Fluo project (ANR-20-CE45-0007).
\copyright 2021 IEEE. Personal use of this material is permitted. Permission from IEEE must be obtained for all other uses, in any current or future media, including reprinting/republishing this material for advertising or promotional purposes, creating new collective works, for resale or redistribution to servers or lists, or reuse of any copyrighted component of this work in other works}
}

\author{Denis Fortun$^1$, Etienne Baudrier$^1$, Fabian Zwettler$^2$, Markus Sauer$^2$, Sylvain Faisan$^1$\\
\small $^1$ ICube - UMR7357, CNRS, University of Strasbourg\\
\small $^2$ Lehrstuhl f\"{u}r Biotechnologie und Biophysik, Biozentrum, Universit\"{a}t W\"{u}rzburg, Germany}

\date{}

\maketitle

\begin{abstract}
In this paper, we address the problem of registering multiple point clouds corrupted with high anisotropic localization noise. Our approach follows the widely used framework of Gaussian mixture model (GMM) reconstruction with an expectation-maximization (EM) algorithm. Existing methods are based on an implicit assumption of spatially invariant isotropic Gaussian noise. However, this assumption is violated in practice in applications such as single molecule localization microscopy (SMLM). To address this issue, we propose to introduce an explicit localization noise model   that decouples shape modeling with the  GMM from noise handling. We use this model for multiview point cloud registration by designing an EM algorithm that considers noise-free data as a latent variable, with closed-form solutions at each EM step. The first advantage of our approach is to handle spatially varying and anisotropic Gaussian noise. The second advantage is to leverage the explicit noise model to impose prior knowledge about the noise that may be available from physical sensors. We show on various simulated data that our noise handling strategy improves significantly the robustness to high levels of anisotropic noise. We also demonstrate the performance of our method on real SMLM data.
\end{abstract}

\section{Introduction}
\label{sec:intro}

The acquisition of point cloud data has become ubiquitous for the perception of the 3D environment with the advent of sensors such as LiDAR and time-of-flight cameras, used for many applications in robotics, autonomous driving or augmented reality. The registration of point clouds is  a fundamental step to achieve various tasks, including 3D scene reconstruction, pose estimation or localization.
Besides these classical application fields, we particularly focus in this work on point cloud data acquired in single molecule localization microscopy (SMLM) \cite{Sauer13}, which is one of the reference techniques in fluorescence microscopy.
Each point in SMLM represents the localization of a fluorophore attached to specific proteins, with a localization accuracy of a few nanometers. 
Registration plays an important role in SMLM in the {\it single particle reconstruction} paradigm \cite{Fortun18}, where  multiple randomly oriented copies of a biological particle are registered in order to reconstruct a single 3D model with  improved resolution.

Most registration methods are based on an iterative procedure that alternates between the computation of distances between points, and the estimation of a rigid transformation that aligns the point clouds. 
The iterative closest point (ICP) method is the oldest and most popular representative of this approach \cite{Besl92}. It retains only the smallest point-to-point distances to establish binary correspondences that enable a fast estimation of the optimal transformation parameters. However, the choice of binary correspondences limits the robustness of ICP to noise, missing data and initialization. 

To circumvent this issue, the binary matches can be replaced by fuzzy correspondences. This generalization of ICP can be formalized by representing the target point cloud as a Gaussian mixture model (GMM) and solving a maximum likelihood clustering problem with an expectation-maximization algorithm (EM) \cite{Gold98,Chui00,Granger02}. In this setting, each point of the target point cloud is the center of a Gaussian component of the GMM. Applying an EM algorithm amounts to maximize an auxiliary function where each point correspondence is weighted by the probability to cluster the point of the source point cloud in the GMM centered on the point of the target. The E-step is the computation of these clustering probabilities, and the M-step is the estimation of the rigid transformation.
This strategy can be generalized further by considering a generative approach \cite{Horaud10,Eckart18}, where the GMM represents the distribution that generated both point clouds. In this framework, the GMM is estimated jointly with the rigid transformation. This is achieved with a conditional EM algorithm: the M-step is subdived in two independent steps, dedicated to the estimation of the rigid transformation and the estimation of the GMM parameters.
The generative approach can be naturally extended to the registration of multiple point sets, and it can reduce considerably the computational complexity when a low number of GMM components is used. A large proportion of multiview registration methods can be cast in this generative framework, which we will call EM-GMM in the remainder of the paper. The EM procedure converges to a local maximum, but global convergence is not guaranteed. Therefore, it is usually used as a refinement, after a coarse alignment of the point clouds.

\begin{figure}[t]
  \centering
  $\begin{array}{m{22pt}@{\hspace{0pt}}m{120pt}@{\hspace{0pt}}m{120pt}}
  \centering Top view &
  \centering\includegraphics[width=70pt]{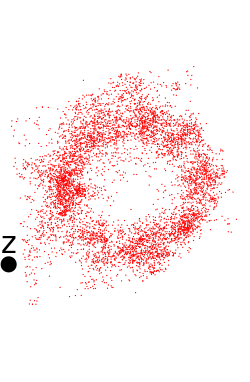} &
  \includegraphics[width=70pt]{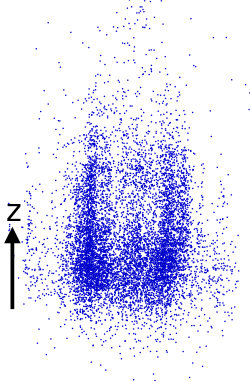} \\[-10pt]
  &\centering\mbox{(a)} & ~~~~~~~~~~~\mbox{(b)} \\
  \centering Side view&
  \centering \includegraphics[width=70pt]{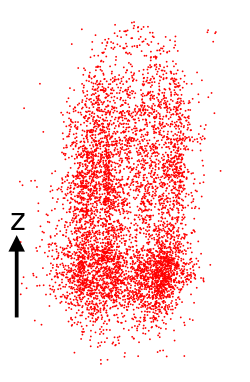} &
  \includegraphics[width=70pt]{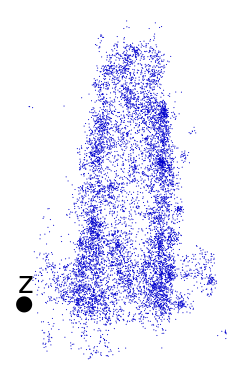}\\[-10pt]
  &\centering\mbox{(c)} & ~~~~~~~~~~~\mbox{(d)} 
  \end{array}$
\caption{Real data of centriole acquired with dStorm \cite{Heilemann08}. The two columns show two examples of point clouds, with a 90 degrees tilt of the centriole between the two views. The first row represent a view in the direction of the axis of symmetry of the centriole (top view), and the second row represents a view parallel to the axis of symmetry (side view).}
\label{fig:real_data}
\end{figure}

A major limitation of EM-GMM methods is the absence of explicit localization noise model.  The localization noise comes from the  measurement imperfection inherent to any physical acquisition devices. This is particularly true in SMLM where the points are estimated by a localization algorithm that usually provides known Gaussian uncertainty associated to each point, with a much higher variance in the axial direction of the microscope. To illustrate the amount of anisotropic noise in SMLM, we show in Figure~\ref{fig:real_data} two acquisitions of a macromolecular assembly called centriole, which is known to have a barrel-like structure. The symmetry axis of the centriole is aligned with the axial direction z in the first acquisition (Figure~\ref{fig:real_data} a,c), and it is orthogonal to z in the second acquisition (Figure~\ref{fig:real_data} b,d). Figure \ref{fig:real_data}.a and \ref{fig:real_data}.b both show a view of the centriole in the direction of its symmetry axis, but the ring shape that corresponds to the circumference of the barrel is clearly visible in  Fig \ref{fig:real_data}.a, whereas the axial uncertainty completely blurs the shape in the z direction in Figure~\ref{fig:real_data}.b.
This example shows how the anisotropic noise can drastically modify the geometry of an object. Consequently, in the absence of noise model, two point clouds such as those of Figure~\ref{fig:real_data} cannot be represented by a common GMM, as it is done in the standard EM-GMM approach. The origin of the problem is that  the GMM has to play two different roles: it has to fit both the shape of the object and the noise.
%This strategy is only valid for space-invariant isotropic Gaussian noise, but fails otherwise. 
Moreover, the absence of explicit noise model also prevents from incorporating prior knowledge about the localization uncertainty available in SMLM and potentially in other acquisition devices. 
%\sylvain{(Je pencherais plutôt pour enlever cette dernière phrase. Si on nous demande de parler des méthodes deep learning on pourra le faire mais is on le mentionne rapidement comme ça on on demandera à coup sur plus de details)}%, which leads to try to match point clouds with different shapes.

In this paper, we propose a generalization of EM-GMM registration methods that takes into account localization noise. We assume that the noise follows a Gaussian distribution with known and anisotropic covariance. Moreover, we consider spatially varying covariance, which means that the covariance can be different for each point.
%\df{localization noise. For each point, the localization noise follows a known but arbitrary Gaussian distribution, with each point being associated with its own Gaussian distribution.}% 
%space-varying Gaussian localization noise with arbitrary \beb known\eeb{} covariances.
We explicitly model the observed  data as a noisy measurement of the unknown clean data. %, with an anisotropic Gaussian uncertainty on the localization of the points.
Thus, the GMM is fitted on the clean data to model the shape of the object, and it is decoupled from the noise model that generated the observed data.
Since the clean data is unknown, our strategy is to consider it as a latent variable in the EM algorithm. % REMOVE, which induces an additional marginalization in the expected log likelihood maximized in the EM algorithm. 
We show that with this new model, each step of the EM algorithm can be solved analytically.
%use a stochastic EM approach to approximate the integral in the estimation of the rigid transformation parameters, and we derive analytical solutions at each step of the algorithm. 
We apply our strategy in the framework of the joint registration of multiple point clouds algorithm (JRMPC) \cite{Evangelidis17}, which is a generic EM-GMM method that can be applied to multiple point clouds. Our approach is generic and could also be integrated in the numerous EM-GMM variants of \cite{Evangelidis17}.
We show experimentally on simulated data that our method significantly improves registration and GMM reconstruction compared to the baseline method in the presence of high anisotropic and spatially varying noise. We also demonstrate the performance and interest of our approach on real SMLM data. 

%The remainder of the paper is organized as follows. In Section \ref{sec:previous_work}, we review related works for point clouds registration. In Section \ref{sec:method}, we describe our proposed method. We give implementation details in Section \ref{sec:impl_details}, and finally, we present and discuss experimental results in Section \ref{sec:results}.

\section{Related works}
\label{sec:previous_work}
In this section, we focus on reviewing the main EM-GMM strategies and their relation with noise handling.

The earliest works on the EM-GMM framework were dedicated to pairwise registration \cite{Chui00,Granger02}. The GMM was designed to fit the target point cloud by defining each of its points as the center of a Gaussian component. In this setting, the E-step amounts to the computation of weights of soft assignments, and the M-step is the estimation of the rigid transformation. The variances can be embedded in a coarse-to-fine scheme \cite{Granger02}, or they can be estimated in the M-step \cite{Myronenko10,Horaud10}. Several variants have been proposed to handle outliers \cite{Hermans11}, topological constraints \cite{Myronenko10}, symmetrisation of the registration \cite{Combes20}, or non rigid deformations \cite{Chui00}.  This approach has been recently extended to the registration of multiple point clouds in \cite{Zhu20}, where a point is modeled as a mixture of Gaussians centered in the nearest neighbor points in each other point clouds.

From the point of view of noise modeling, this formulation is based on an assumption of Gaussian localization uncertainty of each point \cite{Granger02}. When the Gaussians have fixed isotropic covariances, the M-step has an analytical solution and it boils down to standard ICP with fuzzy correspondences. Some methods have extended this approach to handle anisotropic noise \cite{Estepar04,Maier11}. The estimation of the transformation (M-step) becomes a generalized total least squares problem that is solved with an iterative algorithm \cite{Ohta98,Balachandran09}. Leveraging anisotropic covariances has also been proposed for the matching (E-step) in \cite{Maier11}.
%\cite{Horaud10}: SDP relaxation

One of the main limitations of these methods is to rely on the assumption of perfect correspondences between noise-free points in the two point clouds. In practice, physical measurement devices are likely to produce different sampling schemes for each new acquisition. Consequently, the perfect correspondence assumption is violated in most cases. A proper noise modeling should decouple the localization uncertainty from the sampling uncertainty.

Moreover, this category of methods has two major limitations already mentioned in Section~\ref{sec:intro}: it is restricted to pairwise registration (excluding \cite{Zhu20}), and since there are as many Gaussian components as points, the computational complexity can become prohibitive. The generative EM-GMM methods \cite{Eckart15,Evangelidis17} have been developed to address these issues. They estimate the GMM as the common distribution of all the point clouds registered in the same pose. This is done by extending the M-step with the estimation of the means and covariances of the GMM in addition to the transformation parameters. It can be formalized as a conditional EM procedure \cite{Evangelidis17}.

In order to improve the robustness of the baseline approach \cite{Evangelidis17}, some works extended it  to handle data composed of both point localization and orientation \cite{Billings15,Ravikumar17,Min20} with a Hybrid mixture model. %for which EM steps are derived similarly to the  GMM case. 
Modeling outlier points and missing data is also a recurrent issue that is usually addressed with an additional uniformly distributed class \cite{Chui00,Myronenko10,Evangelidis17}, or replacing the GMM distribution with a Student's t-mixture model \cite{Ravikumar18}.

The generative methods implicitly assume that the localization noise in the data is Gaussian and spatially invariant. The vast majority of methods consider isotropic covariance of the Gaussian components. An anisotropic variant can be found in \cite{Eckart15}, but the critical step of estimation of the transformation parameters in the M-step is performed with an isotropic approximation of the covariance, so the anisotropy is not properly handled throughout the algorithm. A solution for the M-step is proposed in \cite{Eckart18} with a linear approximation valid only for small angles. Moreover, as mentioned in Section \ref{sec:intro}, prior knowledge about the noise covariance cannot be included in the model since it is implicitly contained in the covariances of the estimated Gaussian components. 

Note that recent methods based on deep neural networks \cite{Aoki19,Wang19,Yew20, 9709963} also do not rely on explicit noise modeling. These methods are either based on the computation of local \cite{Wang19,Yew20} or global \cite{Aoki19,9709963} features. However, the anisotropic noise in SMLM can drastically modify (locally and globally) the geometry of an object (see Figure~\ref{fig:real_data}). Consequently, we experimented that these registration algorithms fail to register SMLM data. In \cite{Fan21}, we  proposed a neural network approach dedicated to the special case of cylindrically symmetric shapes. It yields a coarse alignment and cannot be applied to non-symmetric structures. %Note that this coarse registration method will be used to initialize the parameters of the rigid transformations for the registration of real data.

Multiview registration in the context of SMLM data has been investigated in a few works. However, the problem is addressed in a simplified setting. In \cite{Broeken15}, a template is assumed to be available and is used as a target for pairwise registration. In \cite{Salas17}, point clouds data is transformed to produce 2D image projections, which are used to apply standard reconstruction methods developed for tomography with unknown angles in cryo-electron microscopy. The method described in \cite{Heydarian21} uses the motion averaging framework, based on the computation of all pairwise registrations between input point clouds, followed by a global fusion of the transformation parameters. The pairwise registration step is performed with the method described in \cite{Jian10}, which assumes isotropic noise, which is incompatible with the highly anisotropic localization uncertainty of SMLM data. Finally, in \cite{Wang22}, the registration method proposed in \cite{Evangelidis17} is applied to SMLM data, combined with a multistart strategy. However, the noise is also assumed to be isotropic and shift invariant and does not fit with the SMLM model.

We also mention that the derivation of a GMM clustering method in the presence of uncertainty in the data has been introduced in \cite{Ozerov13} in the context of acoustic data. %Our work is inspired from the solution of \cite{Ozerov13} for the GMM fitting part of our algorithm.  

%\todo{deep learning methods, motion averaging method?}

%\section{Registration with anisotropic uncertainty modelling}

\section{Generative registration formulation with explicit noise model}
\label{sec:model}

\subsection{Generative model}

%\begin{figure}[t]
%\centering
%  \begin{tabular}{c@{\hspace{40pt}}c}
% \includegraphics[height=130pt]{figures/fig_model_1} &
%  \includegraphics[height=130pt]{figures/fig_model_2}  \\[5pt]
%  \mbox{(a) Standard model} & \mbox{(b) Our model}
%  \end{tabular}
%\caption{Graphical representation of the standard generative model (a), and our generative model with explicit noise handling (b).}
%\label{fig:model}
%\end{figure}

\begin{figure}[t]
\centering
  \begin{tabular}{c@{\hspace{10pt}}c@{\hspace{10pt}}c}
 \includegraphics[height=150pt]{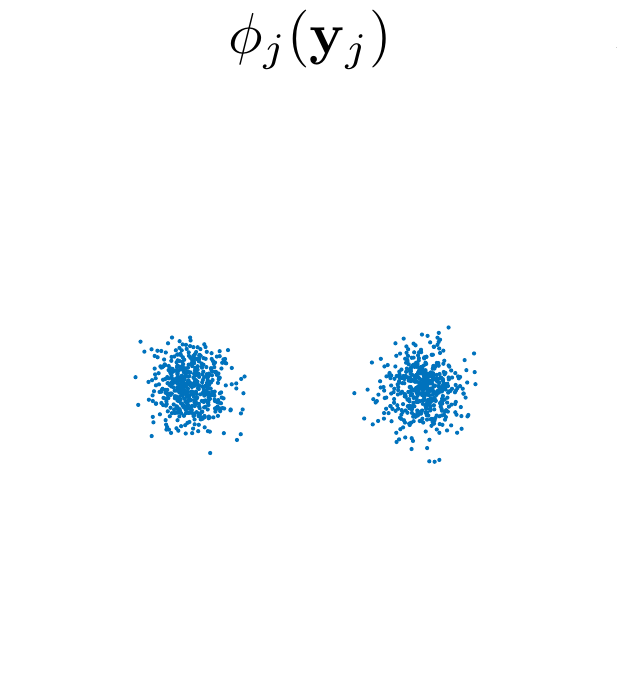} &
 \includegraphics[height=150pt]{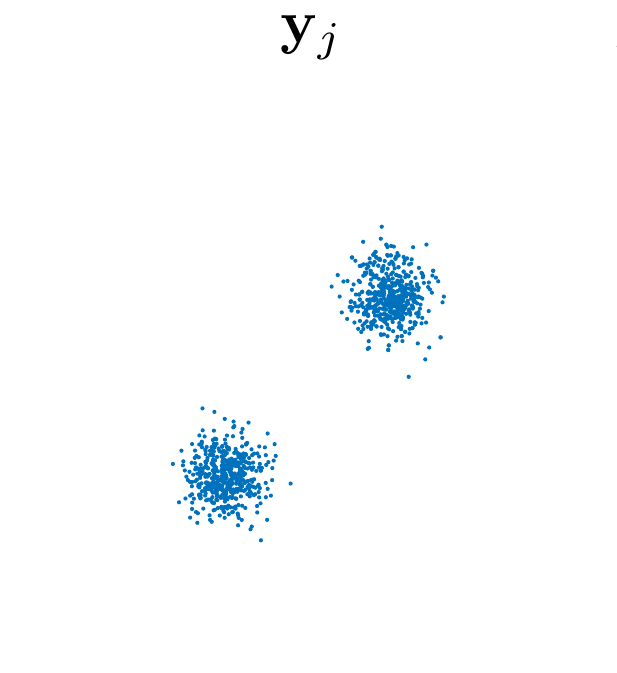} &
 \includegraphics[height=150pt]{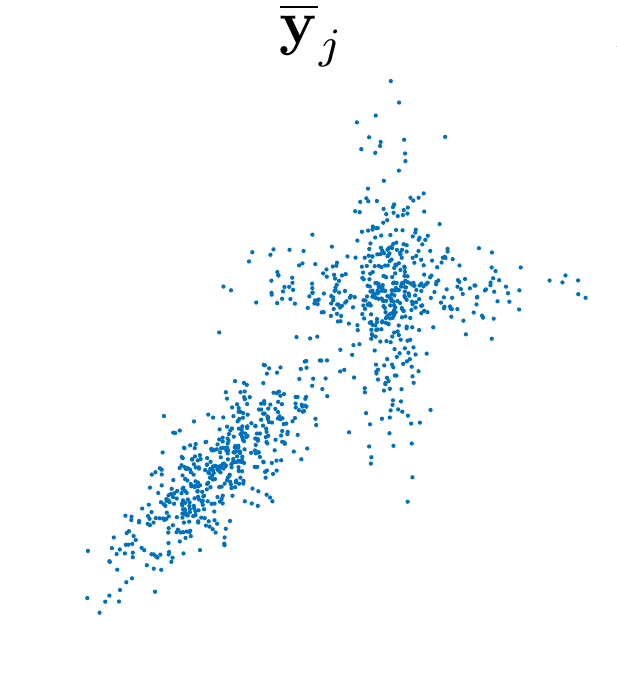} \\
  \mbox{(a)} & \mbox{(b)} & \mbox{(c)}
  \end{tabular}
\caption{% L ancienne version est en commentaire à la fin.
Illustration of our generative model for the synthesis of a point cloud $\by_j$. It consists in  sampling  $\yji$ from (\ref{eq:p_yji}) (steps (a) and (b)) and then  sampling $\byji$  from (\ref{eq:p(byji|yji)}) (step (c)). (a) Sampling from the GMM distribution defined by the parameters $\{p_k,\bmu_k,\bSigma_k\}_{k=1}^K$ used in \eqref{eq:p_yji}.
%(a) Each point of the noise-free point cloud is generated by sampling from the GMM distribution $p(\byji |\theta)$ (Eq. \ref{eq:p_yji}).  
The GMM is here composed of two well separated Gaussians. The obtained points can be denoted as $\phi_j(\y_{j})$ since they correspond to registered points. (b) Noise-free point cloud $\y_{j}$ derived by applying the transformation $\phi_j^{-1}$ to $\phi_j(\y_{j})$.  (c) Observed point cloud $\overline{\y}_j$  obtained as a noisy version of (b) by sampling from $p(\byji|\yji)$ (\ref{eq:p(byji|yji)}). The distribution of the noise may be anisotropic and different for 
each point. Here, the covariances $\bSigmaji$ have elliptic shapes with 
three different orientations clearly visible in the figure: the points that belong to the lower left gaussian component have the same covariance, and the points that belong to the upper right gaussian component are divided into two groups with orthogonally oriented covariances. In the standard GMM model \cite{Evangelidis17}, the generative model only consists in step (a) and (b), whereas we introduce an explicit noise model in step (c).}
\label{fig:model}
\end{figure}

%We have $M$ point clouds \beb and the $j$th is\eeb{}
 We have  $M$ point clouds and the $j$th one is 
denoted by $\overline{\y}_j=(\overline{\y}_{j1},\ldots,\overline{\y}_{jN_j})\in\mathbb R^{3\times N_j}$, where $N_j$ is its number of points% of the $j$th point cloud
. The total number of points is denoted $N = \sum_{j=1}^M N_j$. The goal is to estimate a rigid transformation  composed of a rotation matrix $\R_j\in\mathbb R^{3\times 3}$ and a translation vector $\t_j\in\mathbb R^3$ for each point cloud such that the $M$ transformed points clouds $\R_j \overline{\y}_j + \t_j$ with $j = 1, \ldots, M$  are in the same pose. 

In the generative formulation, all the points of the aligned point clouds are assumed to be drawn from the same distribution, modeled as a GMM  composed of $K$ Gaussian components. Thus, the GMM acts as a soft clustering of the union of all registered point clouds. 
Our purpose is to introduce in this model an explicit noise prior.
To this end, we model the observed points clouds $\by = \{\by_{j} \}_{j=1,\ldots,M} $ as noisy versions of unknown clean point clouds $\y= \{\y_{j} \}_{j=1, \ldots,M}$% \beb and we suppose this noise to be possibly anisotrope and space-varying.\eeb{} 
. Therefore, the distribution of each point $\byji$ is an independent Gaussian perturbation from the clean point $\yji$:
\begin{equation}
p(\byji|\yji) = \mN(\byji;\yji,\bSigmaji), 
\label{eq:p(byji|yji)}
\end{equation}
where $\bSigmaji$ is the known covariance matrix of the measurement noise and 
\begin{equation}
\mN(\x;\bmu,\bSigma)=\frac{1}{\sqrt{(2\pi)^3|\bSigma|}}\exp\left(-\frac{1}{2}\|\x-\bmu\|^2_{\bSigma}\right), 
\end{equation}
with $\|\x-\bmu\|^2_{\bSigma} = (\x-\bmu)^T\bSigma^{-1}(\x-\bmu)$. Note that each point may have a different covariance matrix $\bSigmaji$, and
we do not make any assumption about the form of $\bSigmaji$. 
%\beb Note that each point is associated with its own covariance matrix $\bSigmaji$, with no requirement for it to be diagonal.  \eeb{} % "has a different": plus précis que "is associated with its own"; "with no requirement for it to be": anglais bizarre; 

The distribution of the unknown noise-free data is represented by the GMM
\begin{equation}
\label{eq:p_yji}
p(\yji|\theta)=\sum_{k=1}^K p_k \mN(\phi_j(\yji);\bmu_k,\bSigma_k),
\end{equation} 
where $\bmu_k\in\mathbb R^3$, $\bSigma_k\in\mathbb R^{3\times 3}$ and $p_k\in\mathbb R^+$ are the mean, covariance matrix and weight of the $k$th GMM component, respectively, $\phi_j(\x)=\R_j\x+\t_j$ is the rigid transformation of the $j$th point cloud,  and  $\theta=\left\{\{\bmu_k,\bSigma_k,p_k\}_{k=1}^K,\{\R_j,\t_j\}_{j=1}^M\right\}$ regroups the  parameters of the rigid transformation and the GMM. We make the assumption that $\bSigma_{k} = \sigma_k^2 \I$, with $\sigma_k\in\mathbb R$. Note that the probability $p(\yji|\theta)$ is well defined as a function of $\phi_j(\yji)$ because $\phi_j$ is an invertible one-to-one mapping with a Jacobian determinant equal to one.

Our generative model is graphically illustrated in Fig. \ref{fig:model}. In the standard GMM model, the observations are the noise-free point clouds $\yji$, directly generated from the GMM distribution (Fig. \ref{fig:model}.b). It means that the  localization noise is implicitly assumed to be captured by the GMM distribution. In our model, we consider the noise-free point clouds $\yji$ as a hidden state, which allows us to  explicitly model the noise in the observed point clouds $\byji$ through the probability $p(\byji|\yji)$  (Fig. \ref{fig:model}.c). The idea is to decouple the noise model from the GMM distribution, which is  only used to model the shape of the noise-free data through $p(\yji|\theta)$. %Since the noise covariances $\bSigmaji$ have arbitrary forms and can be different for each point, our model is well-suited to handle anisotropic and space-varying noise.}

%\beb In the standard case \cite{Evangelidis17}, the GMM implicitly contains the noise on the point localization. %If the GMM covariance is isotropic %(which is required to ensure a reasonable computational cost)
%This standard model is not efficient/dedicated to anisotropic and space-varying noise. 
%We propose in this section to have the noise of each point decoupled from the GMM Gaussian shape model.  This opens the way to take into account anisotropic and spatially varying noise, and it allows us to use prior knowledge of the noise.\eeb{}

\subsection{Likelihood}
\label{sec:likelihood}

Under the independence assumption of each point, the likelihood of the point clouds is defined as
\begin{equation}
\label{eq:L}
L(\theta) = p(\overline{\y}|\theta)=\prod_{j=1}^M \prod_{i=1}^{N_j} p(\overline{\y}_{ji}|\theta).
\end{equation}
To find the expression of $p(\overline{\y}_{ji}|\theta)$, we have to marginalize over the unknown clean data $\yji$:
\begin{equation}
p({\byji}|\theta) = \int_{\yji} p(\byji|\yji) p(\yji|\theta)d\yji.
\end{equation}
We show in Appendix \ref{sec::app1} that it can be written	
\begin{equation}
\label{eq:L2}
p({\byji}|\theta) = \sum_{k=1}^K p_k \mN(\phi_j(\byji);\bmu_k,\bSigma_k+\bSigmaji^{\R_j}),
\end{equation}
where $\bSigmaji^{\R_j}=\R_j\overline\bSigma_{ji}\R_j^T$ is the noise covariance $\bSigmaji$ rotated by $\R_j$.
Thus, the likelihood function writes
\begin{equation}
\label{eq:L3}
L(\theta) =  \prod_{j=1}^M \prod_{i=1}^{N_j}\sum_{k=1}^K p_k\,\mN(\phi_j(\byji);\bmu_k,\bSigma_k+\bSigmaji^{\R_j}).
\end{equation}
Our goal is to estimate the GMM and registration parameters $\theta$ that maximize the likelihood~(\ref{eq:L3}).

Note that different parameters $\theta$ correspond to 
equivalent solutions. They represent the same GMM at different poses, and the estimated transformations are modified accordingly. 
Indeed, we can derive from parameters $\theta=\left\{\{\bmu_k,\bSigma_k,p_k\}_{k=1}^K,\{\R_j,\t_j\}_{j=1}^M\right\}$ an equivalent solution parameterized by $\theta_2=\big\{\{\R \bmu_k + \t,\R \bSigma_k \R^t,p_k\}_{k=1}^K,\{\R \R_j ,\R \t_j + \t\}_{j=1}^M\big\}$, where $\R$ is a rotation matrix and $\t$ is a translation vector. It can be easily shown that $L(\theta)=L(\theta_2)$. 
To resolve this ambiguity, it would be possible, for example, to impose $\phi_1$ to be the identity transformation, but we observed that it has no impact on the quality of the registration.
Another well known source of ambiguity is the invariance of the likelihood to permutations in the labelling of the 
Gaussian mixtures (known as the label switching problem), which leads to $K!$ equivalent solutions (without modifying the transformations). 

%In the standard case without noise modeling (Fig. \ref{fig:model}.a), the covariances of the Gausian components in (\ref{eq:L3}) are $\bSigma_k$

%\beb Our likelihood $L(\theta)$ has the same general form as in the standard case, but the covariance of the Gaussian components is $\bSigma_k+\bSigmaji^{\R_j}$ in (\ref{eq:L3}) instead of  $\bSigma_k$ in the standard case. Therefore our likelihood includes explicitly the noise covariance $\bSigmaji^{\R_j}$. This opens the way to anisotropic and spatially varying noise, and it allows us to use prior knowledge of the noise covariance $\bSigmaji$.

%\eeb{}
%In the standard case \cite{Evangelidis17}, the likelihood has the same general form, but the covariance of the Gaussian components are $\bSigma_k$, instead of $\bSigma_k+\bSigmaji^{\R_j}$ in (\ref{eq:L3}). Thus, the GMM covariance $\bSigma_k$ implicitly contains the noise covariance. If $\bSigma_k$ is isotropic (which is required to ensure a reasonable computational cost), this standard model is restricted to isotropic and shift invariant noise covariance. In our formulation (\ref{eq:L3}), the noise covariance of each point $\bSigmaji^{\R_j}$ is decoupled from the Gaussian clusters covariance $\bSigma_k$. This opens the way to anisotropic and spatially varying noise, and it allows us to use prior knowledge of the noise covariance $\bSigmaji$.

\section{EM algorithm for maximum likelihood estimation}
\label{sec:model2}

The optimization of the likelihood (\ref{eq:L3}) is cumbersome and is achieved with the EM algorithm \cite{Dempster77}, which is a widely used procedure for estimating Gaussian mixture densities \cite{Bilmes98}.
The EM algorithm introduces an auxiliary function $Q(\theta,\theta')$, such that if $Q(\theta,\theta') \geq Q(\theta',\theta')$, then $L(\theta) \geq L(\theta')$. Based on this property, the EM algorithm proceeds by defining a series of models $\{{\theta}^{(1)},{\theta}^{(2)},\ldots\}$ such that ${\theta}^{(l+1)}$ maximizes $Q(\cdot,{\theta}^{(l)})$ and thereby $L({\theta}^{(l+1)}) \geq L({\theta}^{(l)})$. 
In this section, we first derive the auxiliary function $Q$ that corresponds to our model presented in Section \ref{sec:model}, then we describe our strategy of space-alternating generalized EM, and finally we describe how we update the GMM and registration parameters by optimizing $Q$ at each iteration.

\subsection{Derivation of the auxiliary function}
To derive the auxiliary function, the problem is reformulated with the introduction of latent variables. 
In the conventional GMM case \cite{Bilmes98,Evangelidis17}, there is only one latent variable $\z=\{z_{ji}\}_{j\in[1\ldots M],i\in[1\ldots N_j]}$, defined such that the value of $z_{ji}\in[1,\ldots,K]$ indicates the Gaussian component assigned to the point $\yji$. The auxiliary function is then defined as the expected log-likelihood  $E[\log(p(\by,\z|\theta))|\by,\theta^{(l)}]$. %REMOVE, which involves a marginalization over $\z$.

In our model, in addition to the Gaussian component assignment $\z$, we also define the set of noise-free point clouds $\y$ as a latent variable.  Thus, the auxiliary function $Q(\theta,\theta^{(l)})$ 
writes:
%REMOVE is a marginalization over the two latent variables $\z$ and $\y$: 
\begin{equation}
\label{eq:Qdef}
\begin{split}
Q(\theta,\theta^{(l)})&= E[\log(p(\by,\y,\z|\theta))|\by,\theta^{(l)}]\\
&=\sum_\z\int_{\mathbb \y} p(\y,\z|\by,\theta^{(l)}) \log(p(\by,\y,\z|\theta))d\y, 
\end{split}
\end{equation}
where we use the notation $\sum_\z$ to denote the sum over all possible values of $\z$.
%where $\theta^{(l)}$ is the model that has been obtained at iteration $l$. As already written, the purpose of Eq. \ref{eq:Qdef} is to estimate a new model $\theta^{(l+1)}$ that maximizes $Q(.,\theta^{(l)})$.
The term $p(\by,\y,\z|\theta)$ of (\ref{eq:Qdef}) is called the {\it complete-data} likelihood because it contains the latent variables, as opposed to the {\it incomplete-data} likelihood (\ref{eq:L3}). 
In our case, we have 	
\begin{equation}
\begin{split}
p(\by,\y,\z|\theta)&=\prod_{j=1}^M \prod_{i=1}^{N_j} p(\byji|\yji)p(\yji|\zji,\theta)p(\zji|\theta). %\\
%& =\prod_{j=1}^M \prod_{i=1}^{N_j}  p_{\zji}  \mN(\byji;\yji,\bSigmaji)\, \mN(\phi_j(\byji);\bmu_{\zji},\bSigma_{\zji}+\bSigmaji^{\R_j}).\textcolor{red}{??????}
\end{split}
\label{eq:L4}
\end{equation}
Using (\ref{eq:L4}), we show in Appendix \ref{sec:app2} that (\ref{eq:Qdef}) writes
\begin{equation}
\label{eq:Q}
%\begin{split}
Q(\theta,\theta^{(l)})= C-\frac{1}{2}\sum_{jik}\Bigg[  \alphajik \int_{\yji} 
\beta_{jik}  (\yji )   \left(\|\phi_j(\yji)-\bmu_k\|^2_{\bSigma_k}  + \log\left|\bSigma_k\right|  - 2\log(p_k) \right) d\yji\Bigg],
%\end{split}
\end{equation}
where $C$ is a constant, we use the notation
$\sum_{jik} = \sum_{j=1}^M\sum_{i=1}^{N_j}\sum_{k=1}^K$, and 
\begin{equation}
\begin{gathered}
\label{eq:ab}
\beta_{jik}  (\yji )  =p(\yji|\zji=k,\byji,\theta^{(l)}),  \\
 \alphajik    =p(z_{ji}=k|\byji,\theta^{(l)}) 
\end{gathered}
\end{equation}
are the conditional distributions of the latent data.
This result is a generalization of the noise-free case: the auxiliary function of the standard model can be obtained by replacing  in (\ref{eq:Q}) the term $\beta_{jik}  (\yji )$ by $\delta(\yji-\byji)$, where $\delta$ is the Dirac function, which removes the integral over the noise-free data $\y$. The presence of this integral in the maximization of (\ref{eq:Q}) makes the problem more challenging than in the standard case, where a standard EM algorithm can be used \cite{Evangelidis17}.\\

\noindent{\bf Expressions of $\alpha_{jik}$ and $\beta_{jik}$}.~
To lighten the notations in what follows, the parameters of $\theta^{(l)}$ and $\theta$ will be both denoted  $p_k$, $\bmu_k$, $\bSigma_k$, $\R_j$  and $\t_j$, without reference to the iteration number $l$.  

Using (\ref{eq:L2}), the expression of $\alpha_{jik}$ is simply:
\begin{equation}
\alpha_{jik}=\frac{\gamma_{jik}}{\sum_{s=1}^K\gamma_{jis}}, 
\label{eq:alpha}
\end{equation}
where 

\begin{equation}
\gamma_{jik} =p_k\mN(\phi_j(\yji);\bmu_k,\bSigma_k+\bSigmaji^{\R_j}). 
\end{equation}
We detail in Appendix \ref{sec:app3} the derivation of the expression of $\beta_{jik}$:
	\begin{equation}
\label{eq:beta}
\beta_{jik} (\yji) = \mN(\phi_j(\yji);\hat{\y}_{jik},(\I-\W_{jik}) \bSigma_k)
\end{equation}
where 
\begin{gather*}
\W_{jik} = \bSigma_k\left(\bSigma_k+\bSigmaji^{\R_j}\right)^{-1}\\
\hat\y_{jik}=\W_{jik}(\phi_j(\byji)-\bmu_k)+\bmu_k.
\end{gather*}
Based on (\ref{eq:beta}), the point $\hat\y_{jik}$ can be seen as a denoised version of $\phi_j({\byji})$ under the 
hypothesis that the point $\yji$ is assigned to the Gaussian component $k$.

\subsection{Space-alternating generalized EM}
%The only terms of (\ref{eq:Q}) that depend on $\theta^{(l)}$ are $\alpha_{jik}$ and $\beta_{jik}$ (see (\ref{eq:ab})). 
In a conventional EM algorithm, the E-step would consist in computing the conditional distributions $\alpha_{jik}$ and $\beta_{jik}$, and the M-step would be the actual maximization of $Q(\theta,\theta^{(l)})$ defined in (\ref{eq:Q}). 
However, maximizing $Q$ with respect to all the parameters in $\theta$  is here an intractable task. This issue can be circumvented by searching for $\theta$ that only guarantees an increase $Q(\theta,\theta^{(l)}) > Q(\theta^{(l)},\theta^{(l)})$. 
This approach is called Generalized Expectation Maximization (GEM) and ensures that the likelihood increases along the iterations. %Several GEM strategies can be considered.

%In  \cite{Evangelidis17}, an Expectation Conditional Maximization (ECM) algorithm \cite{Meng93} is used. The parameters $\theta$ are splitted in two parts: the GMM parameters $\theta_{G}=\{\bmu_k,\bSigma_k,p_k\}_{k=1}^K$, and the transformation parameters $\theta_{T}=\{\R_j,\t_j\}_{j=1}^M$. Accordingly, the M-step is splitted in subproblems related to each variable:
%\begin{eqnarray}
%\theta_{T}^{(l+1)}    &=& \argmax{\theta_{T} }  Q( \{\theta_{T},\theta_{G}^{(l)} \},\theta^{(l)}). \\ 
%\theta_{G}^{(l+1)} &=& \argmax{\theta_{G} } Q( \{\theta_{T}^{(l+1)},\theta_{G} \},\theta^{(l)}).
%\label{eq:op1jj}
%\end{eqnarray}

In this work, we propose to use a Space-alternating GEM  (SAGE)  algorithm \cite{SAGE}. The parameters $\theta$ are split in two parts: the GMM parameters $\theta_{G}=\{\bmu_k,\bSigma_k,p_k\}_{k=1}^K$, and the transformation parameters $\theta_{T}=\{\R_j,\t_j\}_{j=1}^M$. Accordingly, the M-step is split in subproblems related to each variable.
The SAGE algorithm can be written
\begin{eqnarray}
\label{eq:op1}
\theta_{T}^{(l+1)} &=& \argmax{\theta_{T} } Q( \{\theta_{T},\theta_{G}^{(l)} \},\theta^{(l)}). \\ 
\theta_{G}^{(l+1)} &=& \argmax{\theta_{G} } Q( \{\theta_{T}^{(l+1)},\theta_{G} \},
\{\theta_{T}^{(l+1)},\theta_{G}^{(l)} \}).
\label{eq:op2}
\end{eqnarray}

All EM-GMM registration methods use expected conditional maximization (ECM) instead of SAGE.  ECM replaces the update (\ref{eq:op2}) by $\theta_{G}^{(l+1)} = \text{argmax}_{\theta_{G} } Q( \{\theta_{T}^{(l+1)},\theta_{G} \},\theta^{(l)})$, which does not use the updated transformation parameters $\theta_{T}^{(l+1)}$.
The interest of SAGE is to benefit from the updated value $\theta^{(l+1)}_{T}$ to derive new conditional distributions when computing $\theta_{G}^{(l+1)}$ in (\ref{eq:op2}).
The counterpart is that the SAGE algorithm requires two E-steps per iteration whereas the ECM algorithm requires only one E-step per iteration. 

The next two sections detail how we solve the two subproblems (\ref{eq:op1}) and (\ref{eq:op2}). Without loss of generality and in order not to overload the notations, all parameters related to the second operand of $Q$ will be denoted $\theta^{(l)}$.

\subsection{Update of the transformation parameters}

Each rigid transformation $\phi_j$ can be independently computed by maximizing the auxiliary function (Eq. \ref{eq:Q}).
Under the hypothesis $\bSigma_{k} = \sigma_k^2 \I$, we show in Sec. \ref{app:new} that 
the estimation of the rigid transforms amounts to minimize:
\begin{equation}
\{{\R}^{(l+1)}_j,\t^{(l+1)}_j \}=\argmin{\R_j,\t_j}\sum_{ik} \alphajik \| \phi_j(\hat{\y}^{\text{inv}}_{jik})-\bmu_k\|^2_{\bSigma_k}, 
\label{sdf}
\end{equation}
where $\sum_{ik} = \sum_{i=1}^{N_j}\sum_{k=1}^K$, %and where the sought transformation is denoted $\phi^{\star}_j$. Its rotation matrix and translation vector are respectively denoted  $R^{\star}_j$ and $\t^{\star}_j$. Moreover, in Eq. \ref{sdf},
 and $\hat{\mathbf{y}}_{jik}^{\text{inv}}$ denotes the point $\left[\R_j^{(l)}\right]^{-1} \hat{\y}_{jik} - \t_j^{(l)}$. Thus, $\hat{\mathbf{y}}_{jik}^{\text{inv}}$ can be seen as the  denoised point cloud $\hat\y_{jik}$ transformed by the inverse of the current estimation $\phi_j^{(l)}$. Note also that the term $\alpha_{jik}$ is computed based on  $\phi_j^{(l)}$ in \eqref{eq:alpha}.% obtained in the previous iteration of the EM algorithm (Eq. \ref{eq:alpha} and \ref{eq:alpha2}).

Under the same hypothesis $\bSigma_{k} = \sigma_k^2 \I$, \eqref{sdf} is a classical weighted Procruste problem that admits a closed form solution based on SVD \cite{Sorkine09}.

\begin{algorithm}
\caption{Update of the transformation parameters}
\label{alg1}
\begin{algorithmic} 
\STATE {\bf E-step}: computation of conditional expectations and generation of samples according to $\beta_{jik}$.
\STATE $\gamma_{jik} =p_k\mN(\phi_j(\yji)|\bmu_k,\bSigma_k+\bSigmaji^{\R_j})$, $\alpha_{jik} = \frac{\gamma_{jik}}{\sum_{s=1}^K\gamma_{jis}}$
\STATE $\W_{jik} = \bSigma_k\left(\bSigma_k+\bSigmaji^{\R_j}\right)^{-1}$
\STATE $\hat\y_{jik}=\W_{jik}(\phi_j(\byji)-\bmu_k)+\bmu_k$
%\STATE generate $\hat\y^s_{jik}$ by sampling from $\mN(.|\hat{\y}_{jik},(\I-\W_{jik}) \bSigma_k)$
% \STATE $\y^s_{jik} = \R_j^T(\hat\y_{jik}^s - \t_j)$
\STATE $\hat\y^{inv}_{jik} = \R_j^T(\hat\y_{jik} - \t_j)$ 
\STATE { \bf M-rigid step}: update the rigid transformation parameters
\STATE $\{{\R}^{(l+1)}_j,\t^{(l+1)}_j \}=\argmin{\R_j,\t_j}\sum_{ik} \alphajik \|\phi_j(\hat{\y}^{\text{inv}}_{jik})-\bmu_k\|^2_{\bSigma_k}$
%\STATE $\{\R^{(l+1)}_j,\t^{(l+1)}_j\}=\argmin{\R_j,\t_j}\sum_{jik} \alphajik \| \phi_j(\y^{inv}_{jik})-\bmu_k\|^2_{\bSigma_k}$
\end{algorithmic}
\end{algorithm}

\subsection{Update of the parameters of the GMM}
When the transformation parameters are fixed, the optimization problem (\ref{eq:op2}) amounts to a GMM fitting on the union of the aligned point clouds $\{\phi_j(\byji)\}_{j=1 \ldots M}$. Noise-free GMM fitting has a well known solution \cite{Bilmes98}. 
In our model, we have to deal with the integral over the clean data in (\ref{eq:Q}), which makes the problem non-trivial. We show in Appendix~\ref{sec:app4} that the following update equations can be derived from the expression of $Q$ in (\ref{eq:Q}):
\begin{equation}
p^{(l+1)}_k=\frac{1}{N}\sum_{j=1}^M\sum_{i=1}^{N_j}\alphajik,
\label{eq:pk}
\end{equation}
\begin{equation}
\label{eq:muupdate}
\bmu_k^{(l+1)}=\frac{\sum_{j=1}^M\sum_{i=1}^{N_j}\alphajik E[\phi_j(\y_{ji})]}{\sum_{j=1}^M\sum_{i=1}^{N_j}\alphajik},
\end{equation}
\begin{equation}
\label{eq:sigmaupdate}
\bSigma_k^{(l+1)}=\frac{\sum_{j=1}^M\sum_{i=1}^{N_j}\alphajik 
E[(\phi_j(\yji)-\bmu_k)(\phi_j(\yji)-\bmu_k)^T]}{\sum_{j=1}^M\sum_{i=1}^{N_j}\alphajik},
\end{equation}
with the notation $E[X] = \int_{y} \beta_{jik}(y) X dy$. Remarkably, these updates keep the same general form as in the standard noise-free case. The  difference is that the terms that involve the unknown clean data $\phi_j(\yji)$ are replaced here by their conditional expectation. We show in Appendix~\ref{sec:app4} that these conditional expectations have simple closed forms. This leads to the update steps summarized in Algorithm~\ref{alg2}.
\begin{algorithm}
\caption{Update of the parameters of the GMM}
\label{alg2}

\begin{algorithmic}
\STATE {\bf E-step}: computation of conditional expectations
\STATE $\gamma_{jik} =p_k\mN(\phi_j(\yji)|\bmu_k,\bSigma_k+\bSigmaji^{\R_j})$, $\alpha_{jik}=\frac{\gamma_{jik}}{\sum_{s=1}^K\gamma_{jis}}$
\STATE $\W_{jik} = \bSigma_k\left(\bSigma_k+\bSigmaji^{\R_j}\right)^{-1}$
\STATE $\hat\y_{jik}=\W_{jik}(\phi_j(\byji)-\bmu_k)+\bmu_k$.
\STATE $\hat{\R}_{jik} = \hat{\y}_{ji} \hat{\y}_{ji}^T + (\I-\W_{jik}) \bSigma_k $
\STATE {\bf M-GMM step}: update the GMM parameters
\STATE $p_k^{(l+1)} = \frac{1}{N}\sum_{j=1}^M\sum_{i=1}^{N_j} \alphajik$
\STATE $\bmu_k^{(l+1)}=\frac{\sum_{j=1}^M\sum_{i=1}^{N_j}\alphajik \hat\y_{jik}}{\sum_{j=1}^M\sum_{i=1}^{N_j}\alphajik}$
\STATE $\bSigma_k^{(l+1)}=\frac{\sum_{j=1}^M\sum_{i=1}^{N_j}\alphajik \hat{\R}_{jik}}{\sum_{j=1}^M\sum_{i=1}^{N_j}\alphajik}$
\end{algorithmic}
\end{algorithm}

The same algorithm has previously been derived in \cite{Ozerov13} for GMM fitting with uncertainties on acoustic data. In \cite{Ozerov13}, the update equations are directly given after showing that the complete data belongs to the exponential family. We follow here an alternative derivation from the auxiliary function $Q$ with a detailed proof.

\section{Implementation details}
\label{sec:impl_details}

In practice, we do not use the update equation of the priors $p_k$ (\ref{eq:pk})%, in accordance with \cite{Evangelidis17}
. We observed that keeping them constant does not alter the convergence and the quality of the registration.

The method described in Section \ref{sec:model} and \ref{sec:model2} is not robust to outlier points that may not fit with the reconstructed GMM distribution. To overcome this issue, we use the standard approach that consists in adding a class dedicated to outliers, modeled by a uniform distribution. We did not  include this outlier handling in the description of the method for the sake of clarity. When the outlier class is included, Eq. (\ref{eq:p_yji}) becomes
\begin{equation}
\label{eq:p_yji_outliers}
p(\yji|\theta)=\sum_{k=1}^K p_k \mN(\phi_j(\yji);\bmu_k,\bSigma_k) + p_{k+1} \mathcal{U}(h),
\end{equation} 
where $\mathcal{U}(h)$ is the uniform distribution parameterized by the volume 
of the 3D convex hull encompassing the data.
Since the priors $\{p_k\}_{k=1\ldots K+1}$ are not updated, we can show that this modification only modifies the normalization in the computation of the posteriors $\alpha_{jik}$ in the E-step, such that (\ref{eq:alpha}) becomes (for $k=1\ldots K$): 
\begin{equation}
\alpha_{jik}=\frac{\gamma_{jik}}{\sum_{s=1}^K\gamma_{jis}   + \frac{p_{k+1}}{h}    }, 
\label{eq:alph2}
\end{equation}
The outlier handling role of the normalization in (\ref{eq:alph2}) can be understood intuitively.
Suppose that $\byji$ is an outlier.
Without this additional class, since we ensure $\sum_{k=1}^K \alpha_{jik}=1$, all points 
have a similar influence in the update of the parameters.
With this additional class, since $\byji$ is an outlier, we expect $\gamma_{jik} $
to be much smaller than $\frac{p_{k+1}}{h}$, so that the posterior 
$\alpha_{jik}$ is close to $0$ for $k=1...K$ ($\alpha_{jiK+1}$ is close to 1 since 
$\sum_{k=1}^{K+1} \alpha_{jik}=1$). This means that this point has a little impact on the update of the parameters.

%As in \cite{Evangelidis17}, 
We consider GMM with isotropic covariances,  i.e. $\bSigma_k = \sigma^2_k I$. 
In this case, the update of the transformation parameters in (\ref{sdf}) has a simple closed form solution. Concerning the update of the covariance of the GMM in Alg.~\ref{alg2}, we can easily show that the variance parameter can be obtained by ${\sigma^2_k}^{(l+1)} = \text{trace}(\bSigma_k^{(l+1)})/3$.

\section{Experimental Results}
\label{sec:results}
In this section we provide an experimental analysis of our method on synthetic and real data, including a comparison with the baseline method JRMPC \cite{Evangelidis17}. 
%Since our algorithm is an extension of the baseline method JRMPC \cite{Evangelidis17}, we focus our  evaluation on the comparison with this method.

\subsection{Results on synthetic data}

\subsubsection{Dataset}

\begin{figure}[t]
  \centering
  $\begin{array}{c@{\hspace{0pt}}c}
  \includegraphics[width=100pt]{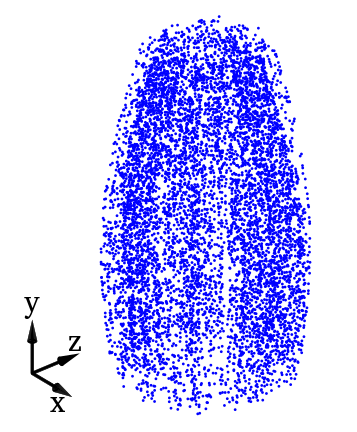} &
  \includegraphics[width=120pt]{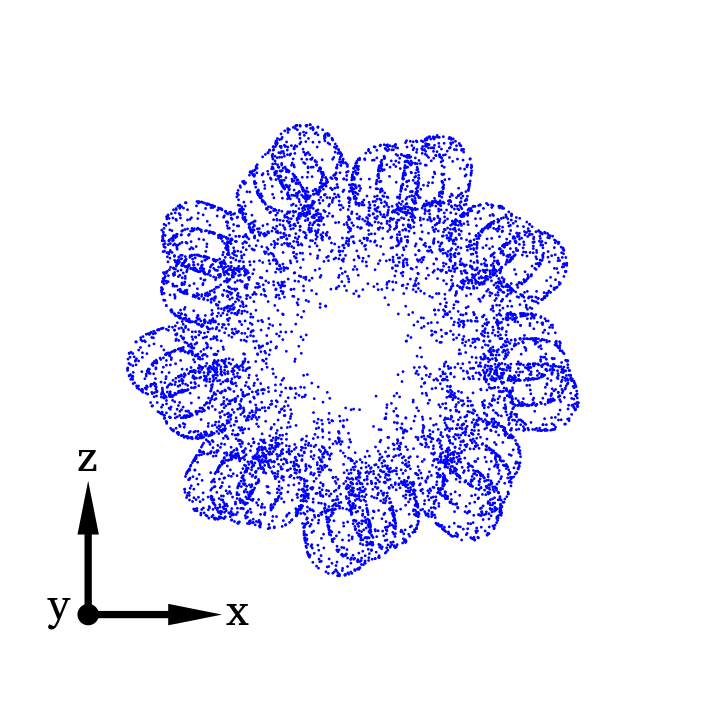} \\
  \includegraphics[width=120pt]{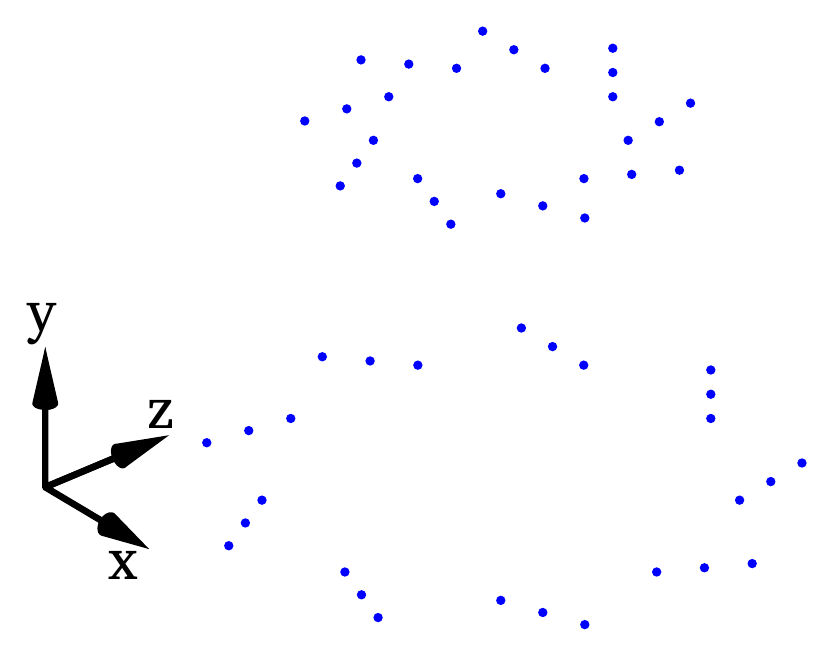} &
  \includegraphics[width=120pt]{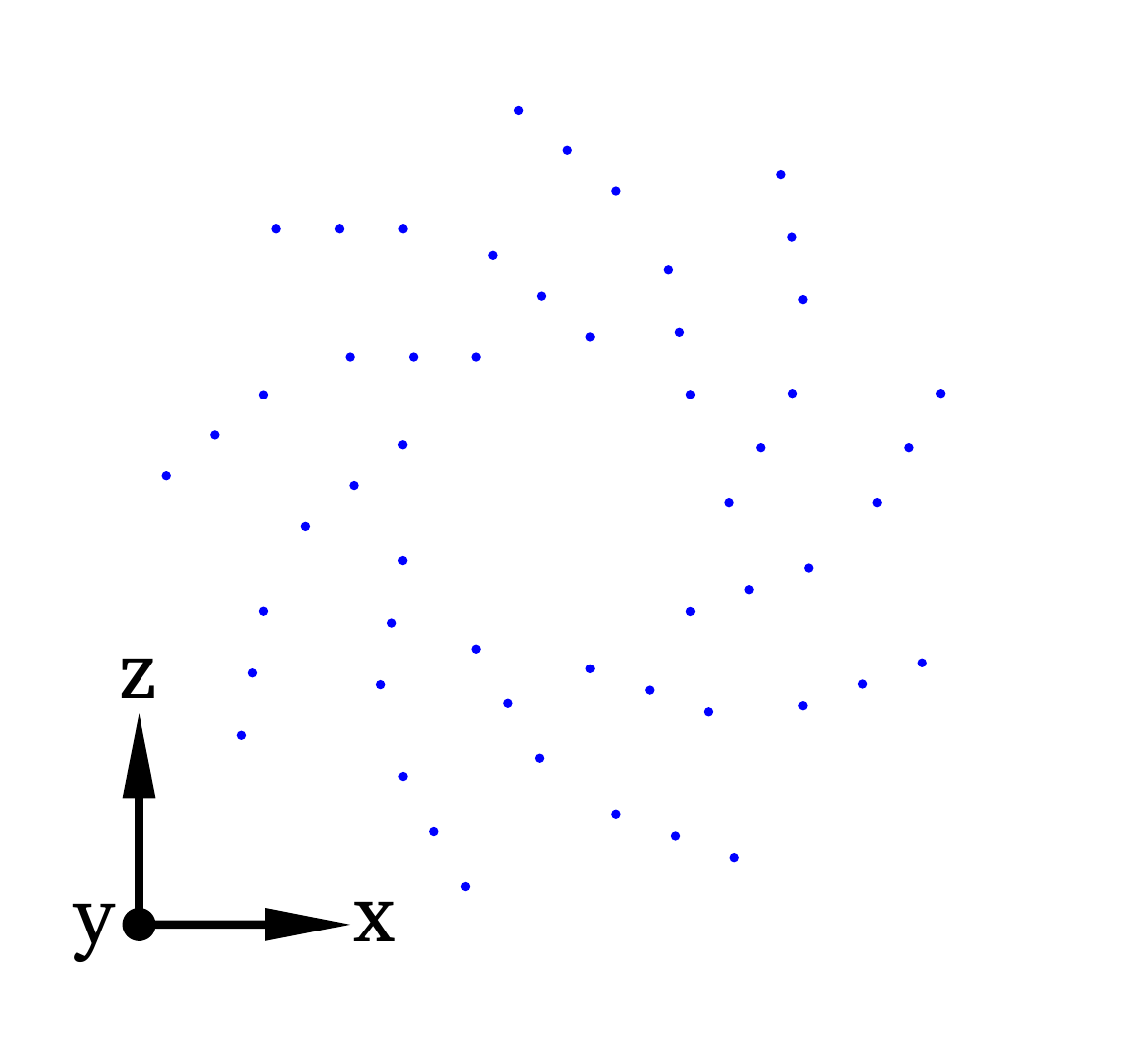} 
  \end{array}$
\caption{Simulated data of the {\it centriole} (top) and {\it triplets} (bottom).}
\label{fig:ground truth}
\end{figure}
\begin{figure*}[t]
  \centering
  $\begin{array}{c@{\hspace{0pt}}c@{\hspace{0pt}}c@{\hspace{0pt}}c}
  \mbox{Ground truth} & \mbox{$r=1$} & \mbox{$r=5$} & \mbox{$r=10$}\\
  \includegraphics[width=108pt]{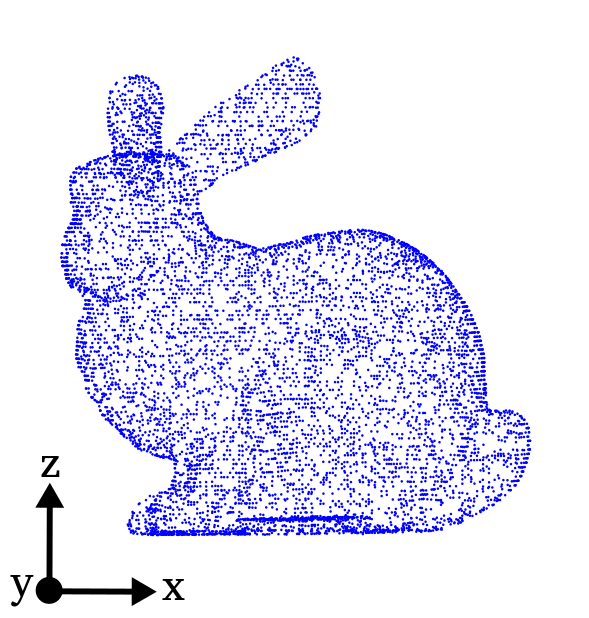} &
  \includegraphics[width=108pt]{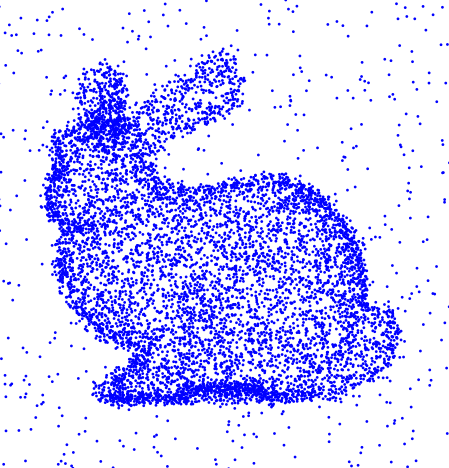} &
  \includegraphics[width=108pt]{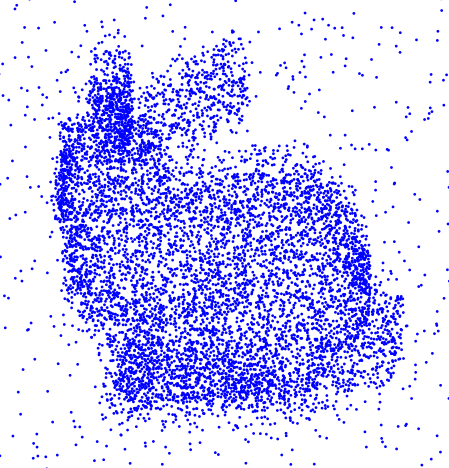} &
  \includegraphics[width=108pt]{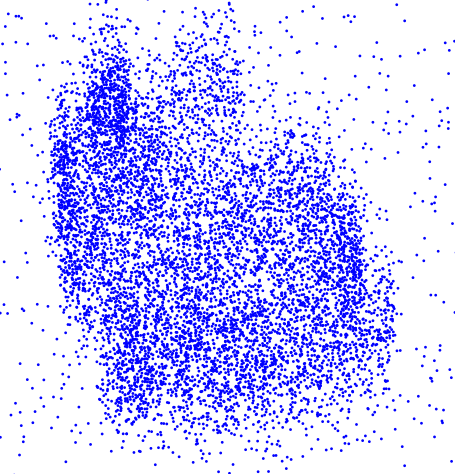} \\
  \includegraphics[width=108pt]{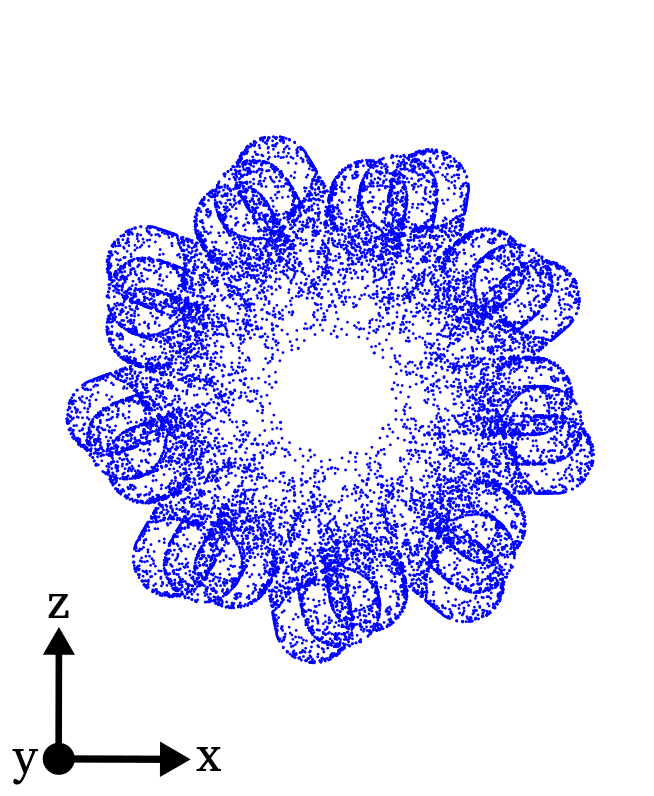} &
  \includegraphics[width=108pt]{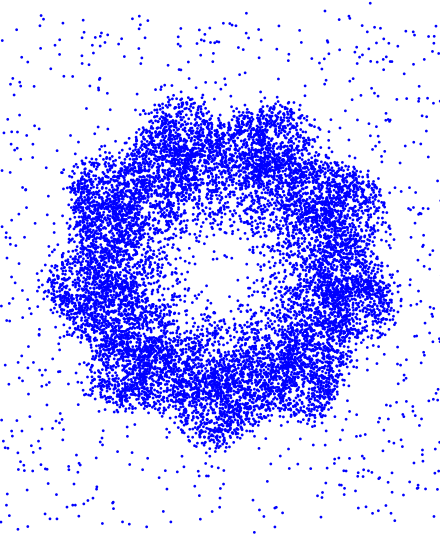} &
  \includegraphics[width=108pt]{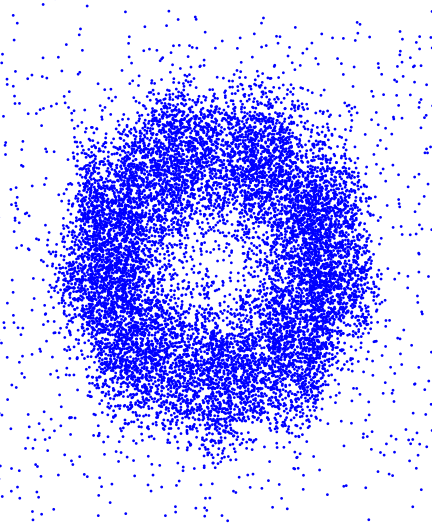} &
  \includegraphics[width=108pt]{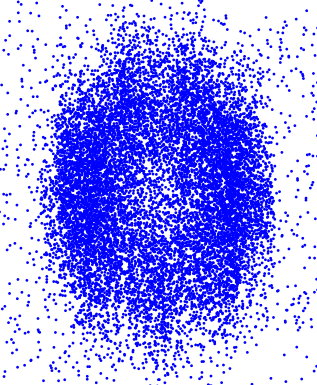} \\
  \end{array}$
\caption{Examples of simulated point clouds with different levels of anisotropic noise from the \textit{bunny} (top) and \textit{centriole} (bottom) models ($\sigma=0.01$ in all the examples). The point clouds are shown in the same pose.}
\label{fig:simulated_data}
\end{figure*}

Since our primary goal is to apply our algorithm to microscopy data, we simulated the tubulin structure of the centriole, which is a fundamental macromolecular assembly involved in most cellular processes. To this end, we used the structural analysis of \cite{LeGuennec20}, where detailed information about the geometry of the centriole is provided.  Our simulated structure is shown in Figure~\ref{fig:ground truth}. It is composed of 9 triplets of overlapping tubes, organized in a ninefold cylindrical symmetry (C9) around the z-axis. It creates a barrel-like structure, with a smaller radius at the top of the structure than at the bottom. We will call this object {\it centriole}.
We also simulated an object composed of 54 source points also organized in triplets with a C9 symmetry (Figure~\ref{fig:ground truth}). It is reminiscent of the structure of the centriole and of other biological particles such as nuclear pore. We will call this object {\it triplets}.
Finally, we also used the {\it bunny} model from the traditional Stanford dataset\footnote{https://graphics.stanford.edu/data/3Dscanrep/}. We fix the number of points of the ground truth model to 2000 for {\it centriole} and {\it bunny}.

A point cloud is created by first rotating the ground truth model, then adding outliers drawn from a uniform law, and finally drawing each point from the Gaussian uncertainty distribution $p(\byji|\yji)$. Our method does not impose a specific form of the covariance matrices of the noise $\bSigmaji$. In our experiments, we  define the covariances as follows:
%Although the proposed approach does not necessitate the covariance matrices $\bSigmaji$ to be diagonal, they have been configured as diagonal matrices for the sake of simplicity. Each matrix $\bSigmaji$ is randomly generated according to Eq. \ref{eq:noise_cov}. The resulting matrices are subsequently assumed to be known during the registration process.
\begin{equation}
\bSigmaji=\left(
\begin{array}{ccc}
\max\left (\sigma +\frac \sigma {10} X_1, 1e^{-5}\right) & 0 & 0\\
0 & \max\left (\sigma +\frac \sigma {10} X_2, 1e^{-5}\right) & 0 \\
0 & 0 & \max\left ( r\sigma +\frac \sigma {10} X_3, 1e^{-5}\right) \\
\end{array}\right)\enspace,
\label{eq:noise_cov}
\end{equation}
where $X_l\sim\mN(0,1)$ ($l=1\ldots 3$), $\sigma > 0 $ is the noise level (variance), and $r \geq 1$ imposes anisotropy in one direction.

%\beb Let us recall that each point measurement is obtained from a clean point by adding a space varying and anisotropic noise:
%$$\byji=\phi^{-1}(\yji)+\varepsilon_{ji},\text{ with } \varepsilon_{ji}\sim\mN(0,\bSigmaji)$$
%We consider uncertainties of the form 
%\begin{equation}
%\bSigmaji=\left(
%\begin{array}{ccc}
%\sigma_{ji}^{(1)} & 0 & 0\\
%0 & \sigma_{ji}^{(2)} & 0 \\
%0 & 0 & \sigma_{ji}^{(3)} \\
%\end{array}\right)\enspace,
%\label{eq:noise_cov}
%\end{equation}
%where, for $1\leq l\leq 3,$ 
%$$\sigma_{ji}^{(l)}\sim\max\left ( r_l\sigma +\frac \sigma {10} X_l, 1e^{-5}\right)\enspace,$$ 
%with $X_l\sim\mN(0,1)$, $\sigma$ the reference variance level and $r_l$ the anisotropy factor in direction $l$. For our numerical application we take $\sigma = 0.01$, $r_1, r_2=1, r_3=10$.\eeb{}
It mimics SMLM data, where the uncertainty in the axial direction is larger than in the lateral plane and it also simulates spatially varying noise since the noise level at each point is defined as a sample of a Gaussian distribution centered on $\sigma$ (respectively $10\sigma$) in the lateral plane (respectively axial direction). In Figure~\ref{fig:simulated_data}, we show examples of simulated point clouds with several values of $r$. %\beb Note that if the covariance matrix is diagonal before the rotation, it is not after the rotation, so this is not a loss of generality.\eeb{} 
We fix the proportion of outliers points to $10\%$ in all our experiments. Finally, the number of point clouds has been set to $M=5$.

\subsubsection{Error metric}

\begin{figure*}[!t]
  \centering
\includegraphics[width=350pt]{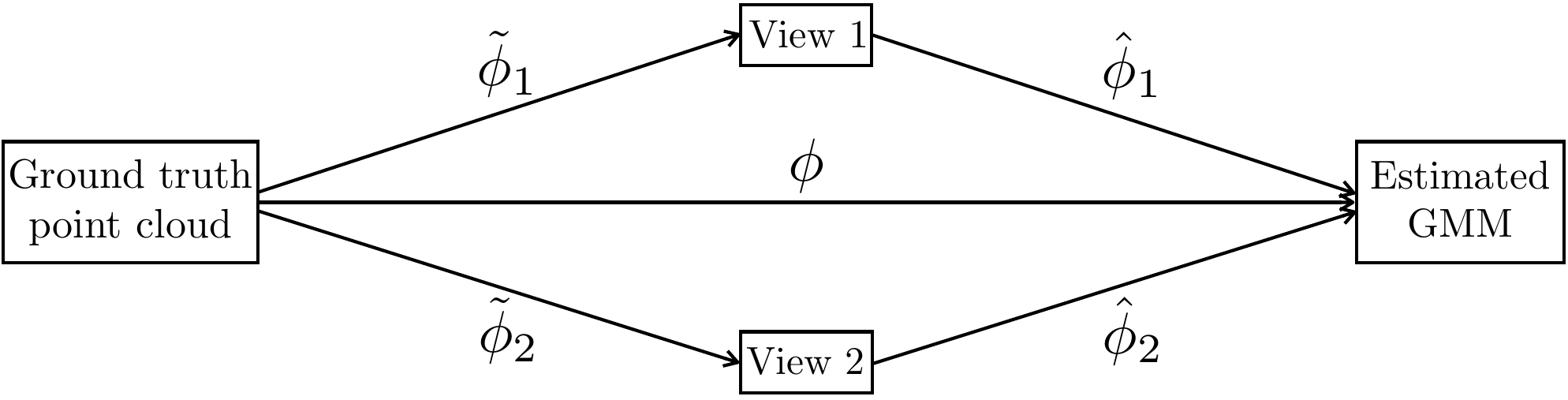} \\ 
\caption{Illustration of the ground truth $\tilde\phi_j$ and estimated $\hat\phi_j$ transformations involved in the computation of our metric, in the simplified case of only two point clouds. $\phi$ represents the transformation between the ground truth and the estimated GMM.}
\label{fig:error}
\end{figure*}

Let us define $\hat{\phi}_j$ as the estimated transformation and $\widetilde{\phi}_j$ as the ground truth tansformation (see an illustration in Figure \ref{fig:error}).
 As explained in the previous section, without considering degradations (noise and outliers), the $i$th observed point cloud is obtained by transforming the ground truth point cloud with the ground truth transformation $\widetilde{\phi}_j$.
Thus, if the estimated GMM were in the same pose as the ground truth point cloud, we would define our error metric as the distance between  $\hat{\phi}_j\circ\widetilde{\phi}_j$ and the identity. However, as pointed in Section \ref{sec:likelihood}, the model is ambiguous because the pose of the estimated GMM is arbitrary. Ideally, we would like to use the  rigid transformation $\phi$ between ground truth and the estimated GMM to compute the distance between $\hat{\phi}_j\circ\tilde{\phi}_j $ and $\phi$ for $j \in [1,M]$. Since $\phi$ is unknown, we define the error metric as the consistency between the estimated transformations of all the views by computing the distances between  ${\hat{\phi}_i} \circ {\tilde{\phi}_i}$ and
${\hat{\phi}_j} \circ  {\tilde{\phi}_j}$, for $(i,j) \in [1,M]^2$ (see Figure \ref{fig:error} for a visual illustration).
%As illustrated in Figure \ref{fig:error}, this distance measures the discrepancy between the registrations of the ground truth model on the estimated GMM when using the ground truth $\tilde \phi_j$ of the views as the first part of the transformation. 
If ${\hat{\phi}_i}\circ {\tilde{\phi}_i}$ is close to  ${\hat{\phi}_j}\circ {\tilde{\phi}_j}$, then ${\hat{\phi}_{ij}} =   {\hat{\phi}_i}^{-1}\circ {\hat{\phi}_j}$ is close to ${\tilde{\phi}_{ij}} = {\tilde{\phi}_i}\circ {\tilde{\phi}_j}^{-1}$. Thus, our error metric is defined as the distance between  ${\hat{\phi}_{ij}}$ and ${\tilde{\phi}_{ij}}$. 

To show that this metric is independent from the pose of the estimated GMM, let us consider a set of estimated transformations $\{\phi_j\}_{j\in[1,M]}$ that correspond to a given pose of the GMM. If the GMM is rigidly transformed by $\phi_0$, the new associated transformations are $\{\phi_0\circ\phi_j\}_{j\in[1,M]}$. The term $\hat\phi_{ij}^0$ associated with these new poses is $\hat\phi_{ij}^0=\hat\phi_i^{-1}\circ\phi_0^{-1}\circ\phi_0\circ\hat\phi_j=\hat\phi_{ij}$. Thus, $\hat\phi_{ij}$ is independent from the pose of the GMM, and so is our error metric.

The translation parameters are easy to estimate (when the rotation parameters are well-estimated), so we only consider the rotation parameters to evaluate the accuracy of the registration. Thus, our error metric is defined as the distance between 
$\hat{R}_{ij}={\hat{R}_i}^T {\hat{R}_j}$ and
$\widetilde{R}_{ij}={\widetilde{R}_i} {\widetilde{R}_j}^T$.

 %XXX Moreover, if ${\hat{R}_i} {\widetilde{R}_i}$ is close to  ${\hat{R}_j} {\widetilde{R}_j}$, then ${\hat{R}_{ij}} =   {\hat{R}_i}^T {\hat{R}_j}$ is close to ${\widetilde{R}_{ij}} = {\widetilde{R}_i} {\widetilde{R}_j}^T$. 
%Thus, our error metric is defined as the distance between 
%${\hat{R}_{ij}}$ and ${\widetilde{R}_{ij}}$.
%Thus, our error metric is defined as the distance between the matrices ${\hat{R}_i} {\widetilde{R}_i}$ and ${\hat{R}_j} {\widetilde{R}_j}$, or equivalently between the matrices ${\hat{R}_{ij}} =   {\hat{R}_i}^T {\hat{R}_j}$ and ${\widetilde{R}_{ij}} = {\widetilde{R}_i} {\widetilde{R}_j}^T$ XXX .
Several  functions have been proposed to measure the distance  between  3D  rotation matrices \cite{Rotation}. 
The distance $d_{ij}$ between $\hat{R}_{ij}$ and $\widetilde{R}_{ij}$ can be computed from the angle of the rotation matrix $ {\hat{R}_{ij}} {\widetilde{R}_{ij}}^T$ (that should be close to the identity matrix). This angle can be computed using the quaternion representation (as in \cite{Rotation}), or simply by:
\begin{equation}
\theta_{ij} = \arccos\left(\frac{\trace({\hat{R}_{ij}} {\widetilde{R}_{ij}}^{T})-1}{2} \right).
\label{eq:toto}
\end{equation}
The distance $d_{ij}$ is then 
\begin{equation}
\label{eq:dij}
d_{ij}=\min(\theta_{ij},\pi-\theta_{ij})
\end{equation}
(it is possible to compute $d_{ij}$ directly from (\ref{eq:toto}) by using an absolute value for the term inside the $acos$ function). 

The {\it centriole} and {\it triplets} models have a ninefold cylindrical symmetry with respect to the z-axis. This means that the rotation by an angle of $2k\pi/9$ ($k=0 \ldots 8$)  does not change the object.
In this case, the error metric has to be adapted since there are nine correct rotation parameters. 
We write $d_{ij}(k)$ the distance defined in (\ref{eq:dij}) between ${\hat{R}_{ij}} =   {\hat{R}_i}^T {\hat{R}_j}$ and ${\widetilde{R}_{ij}}(k) = {\widetilde{R}_i} R_z(k)^T {\widetilde{R}_j}^T$, where $R_z(k)$ defines the rotation matrix around the z-axis by an angle of $2k\pi/9$. We define the error metric by computing the nine possible cases and selecting the minimum value: $d_{ij} = \min_{k =0\ldots 8}d_{ij}(k)$.

\subsubsection{Hyperparameter setting and initialization}
%\beb The determination of the number of Gaussian components $K$ conditions the resolution of the GMM model, and algorithms exist to aid in its selection. Nevertheless, the GMM is utilized in our algorithm for registering the point clouds.\eeb{} %
We observe that $K$ does not have a substantial impact on the registration results as long as it is of the same order of magnitude as the number of points. %\beb Therefore, we fix it beforehand to ensure optimal speed for the registration. \eeb
We set $K = 54$ for the \textit{triplets}, and $K = 2000$ for both the \textit{centriole} and the \textit{bunny}.

As explained in the previous section and as in [13], the priors $p_k$ are not updated and are all set to the same value except $p_{K+1}$ that represents the outlier probability. The proportion of outliers can be defined as
\begin{equation}
\gamma=\frac{p_{K+1}}{\sum_{k=1}^{K}p_k}=\frac{p_{K+1}}{Kp_1}.
\end{equation}
We observed that variations of the value of $\gamma$ between 0.05 and 0.2 had very little impact on the registration results. In the following, the priors $p_k$ have been set so that $\gamma=0.1$. 

Initialization plays a crucial role in EM procedures. %As in \cite{Evangelidis17}, 
We suppose we have a rough initialization of the transformation parameters. Rotation parameters are initialized by drawing the Euler angles from a Gaussian law centered on the ground truth angles, with a standard deviation of $30$ degrees. For the initialization of the GMM, we use these transformation parameters to transform each point set
to the common space and we select randomly $K$ different points among the $M$ transformed points to initialize $\mu_k$. The variances $\sigma_k$  are initialized with high values. 

We use the same hyperparameters and initialization for JRMPC and our method.
To account for the random nature of the initialization, we run both algorithms five times with different GMM initializations and retain for each one the result with the highest likelihood.

\begin{figure*}[t]
  \centering
  $\begin{array}{c@{\hspace{0pt}}c@{\hspace{0pt}}c}
  \mbox{\it{Triplets}} & \mbox{\it{Centriole}} & \mbox{\it{Bunny}} \\ [-1pt]
  \includegraphics[width=130pt]{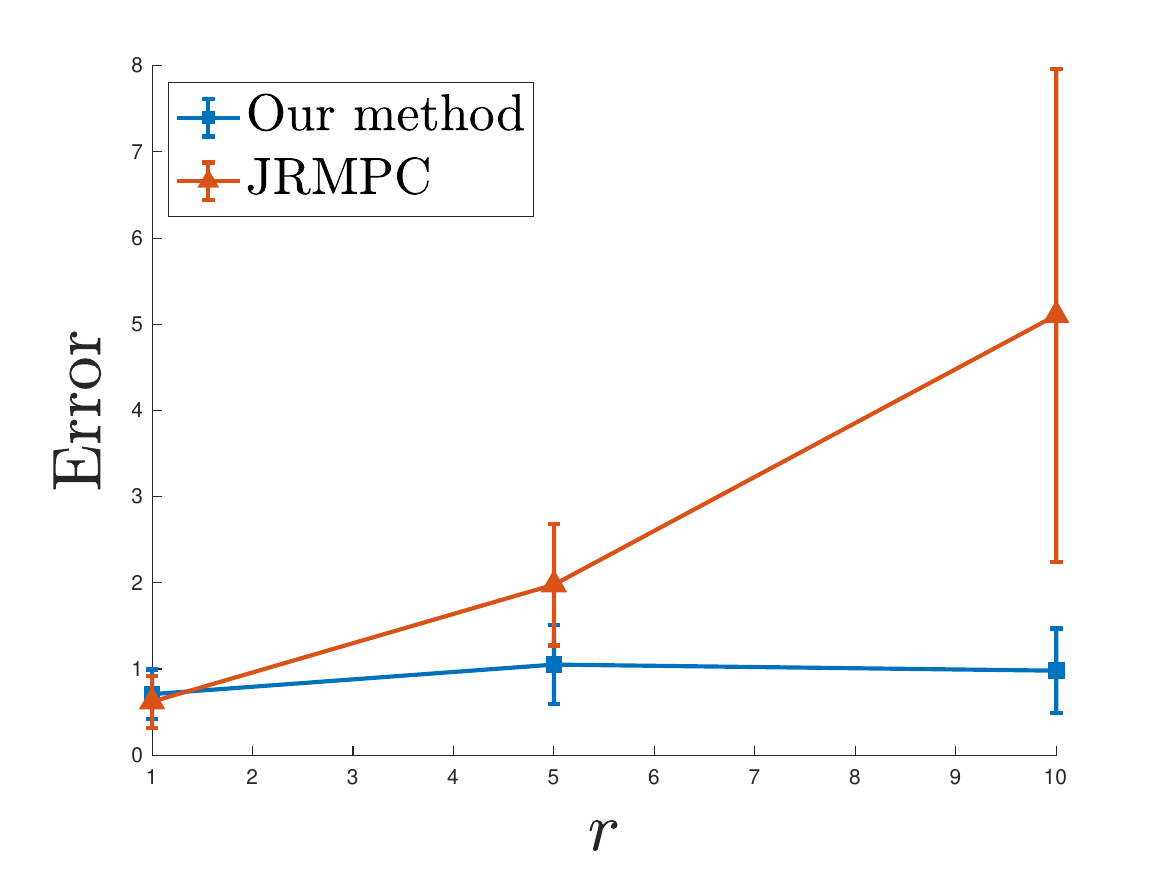} &
  \includegraphics[width=130pt]{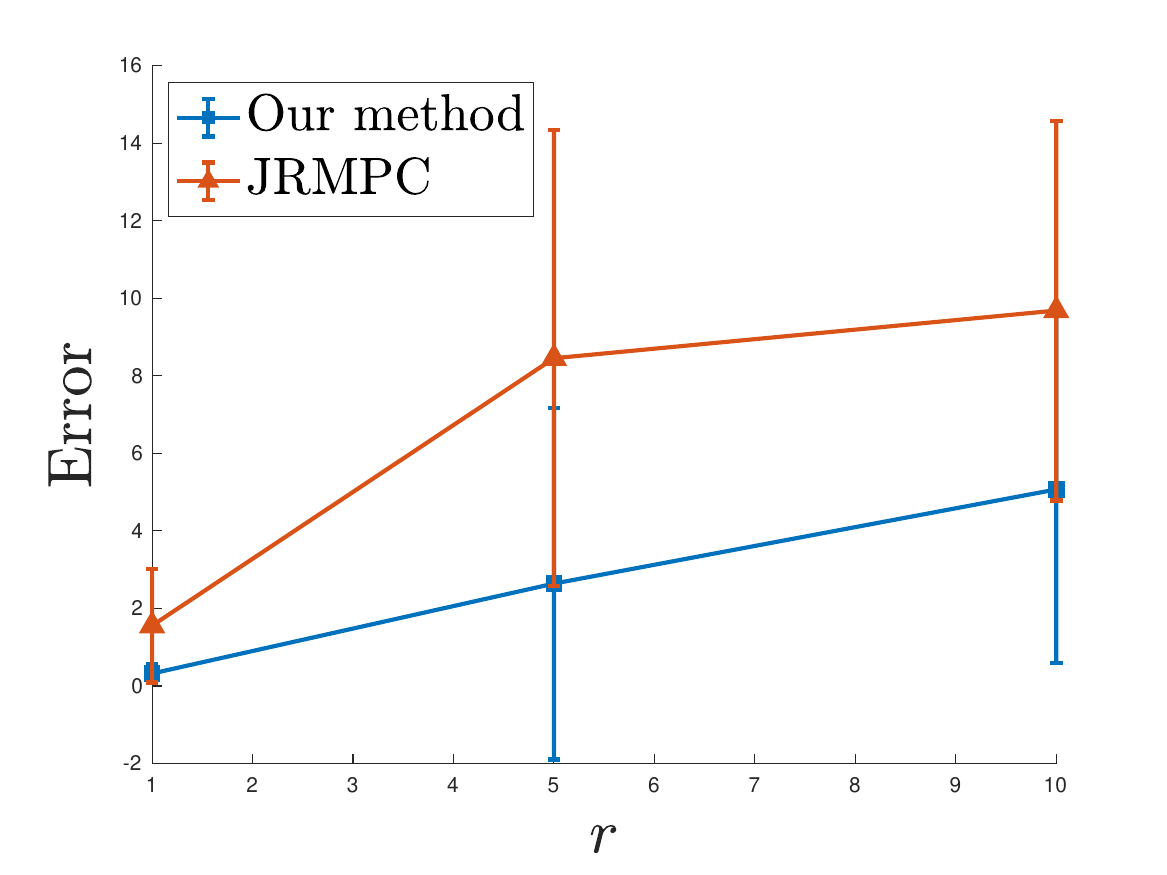} & 
  \includegraphics[width=130pt]{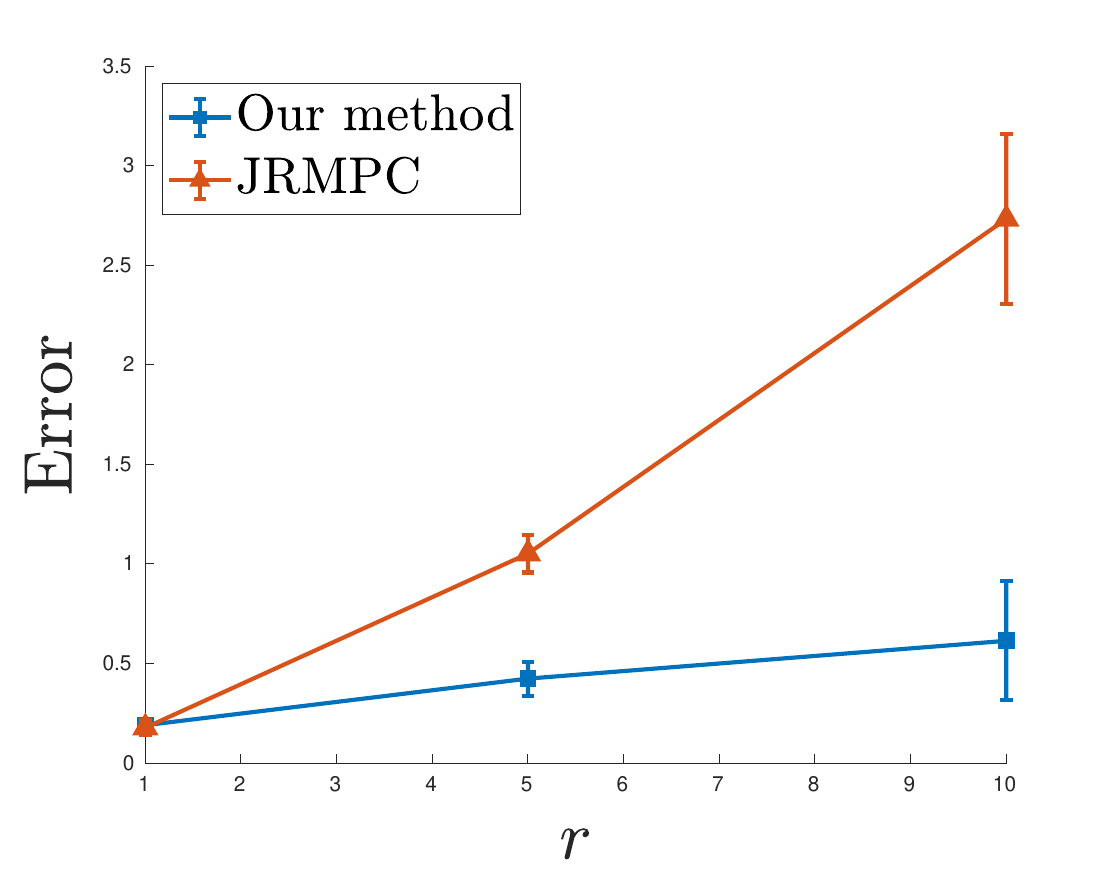} \\ 
  \includegraphics[width=130pt]{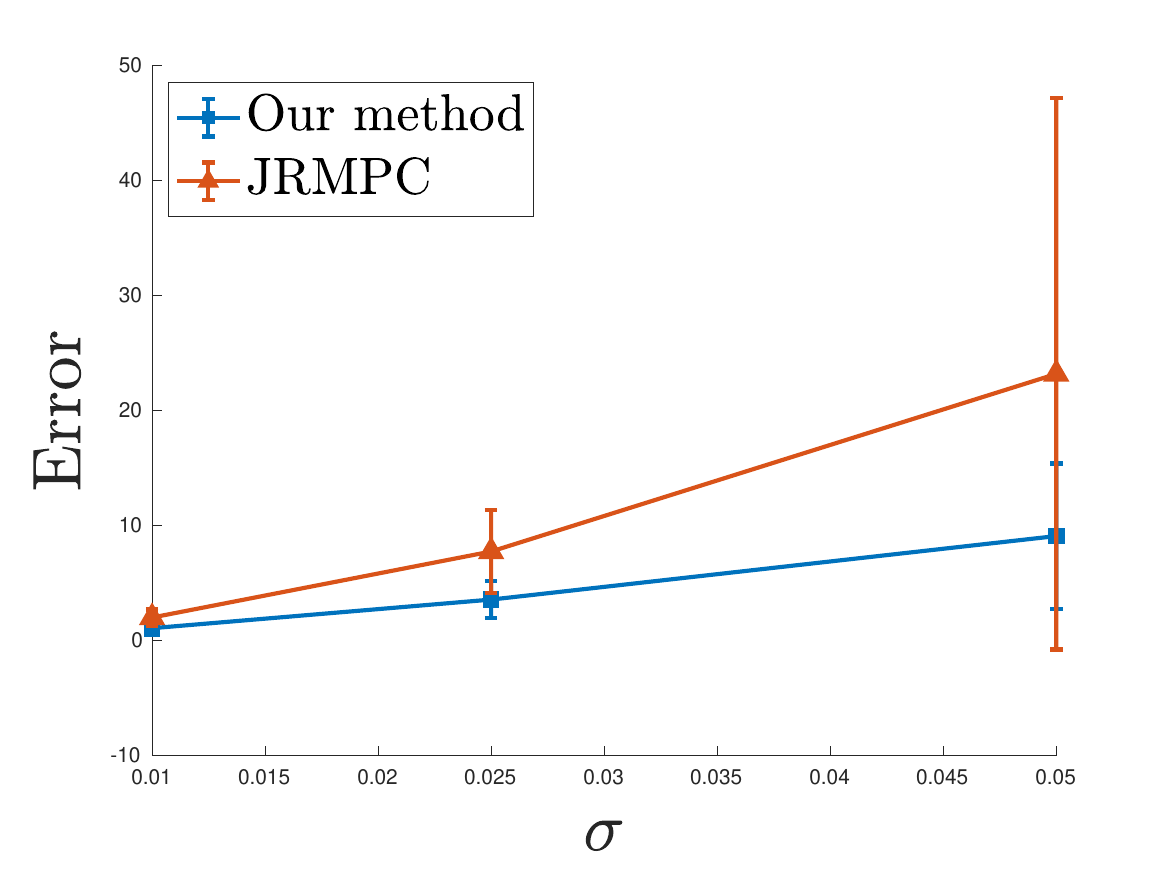} &
  \includegraphics[width=130pt]{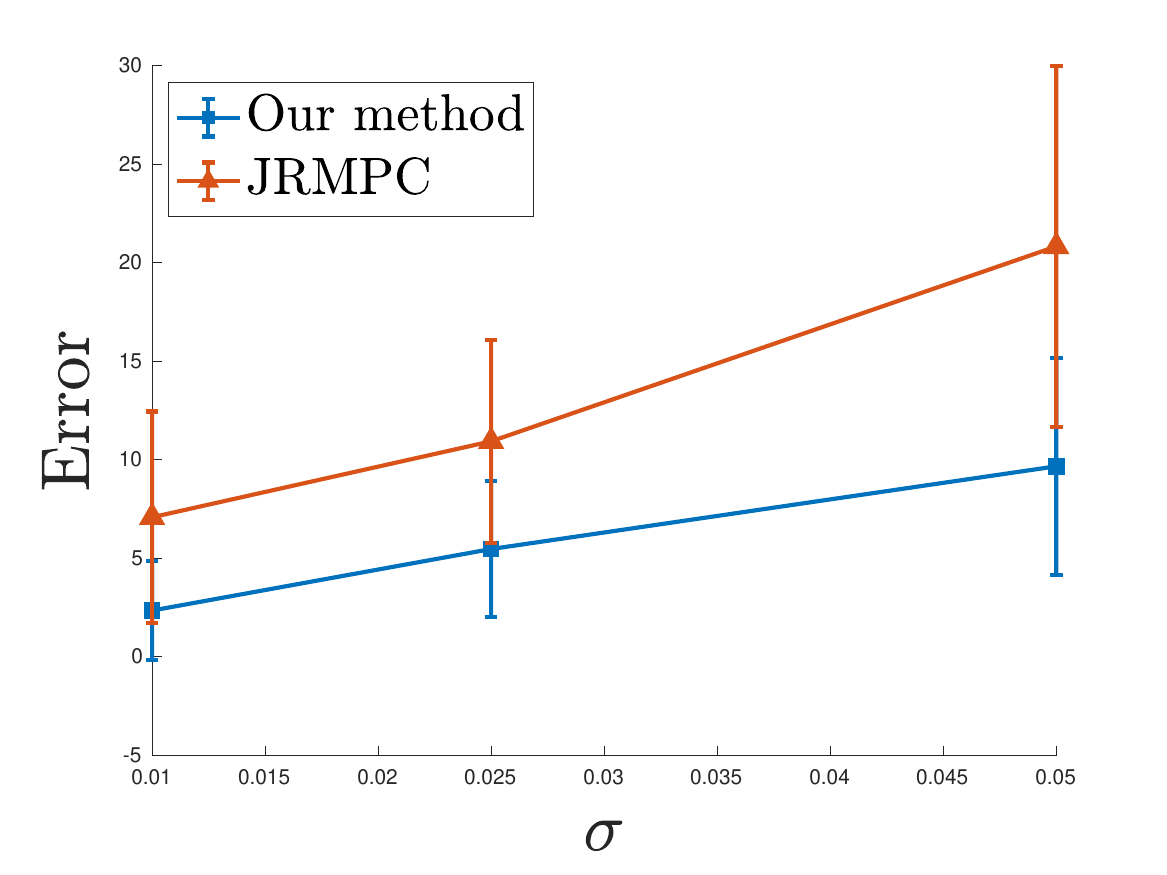} & 
  \includegraphics[width=130pt]{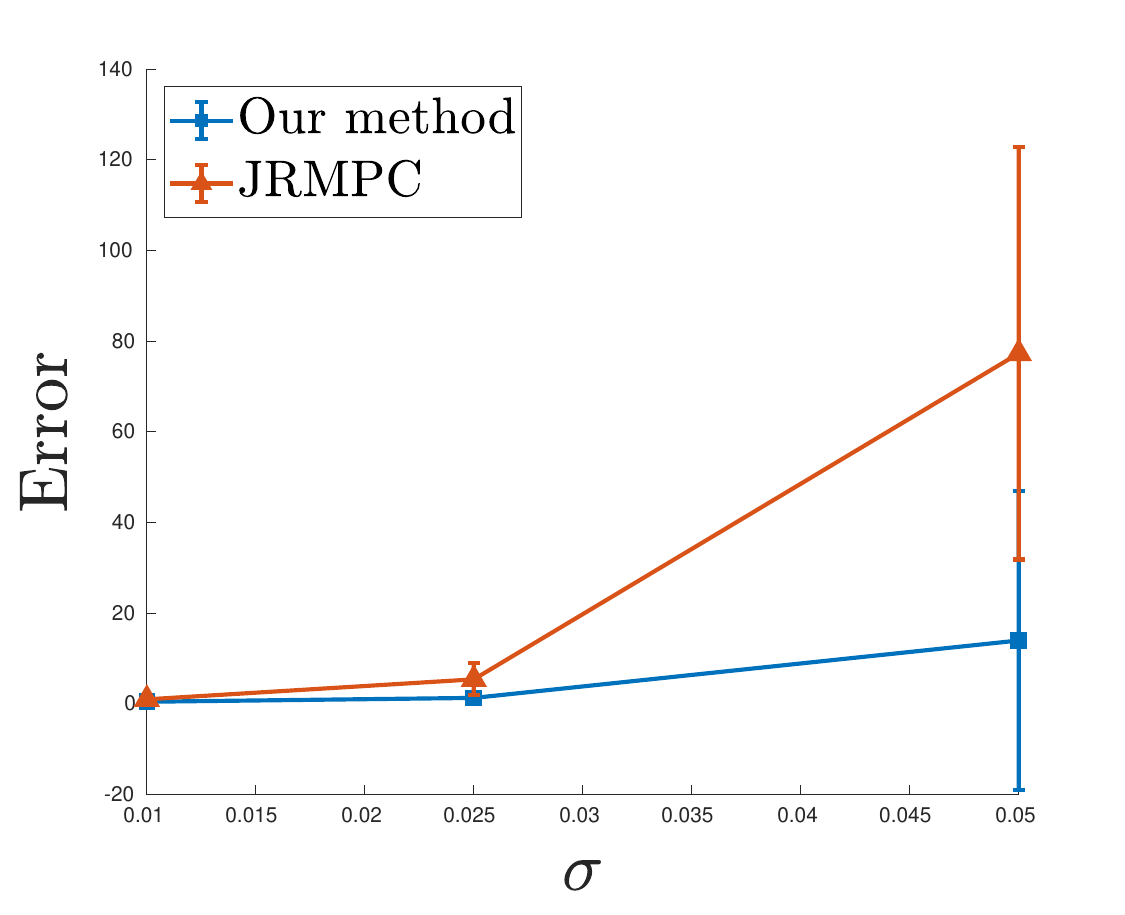} \\ 
  \end{array}$
\caption{Impact of different noise parameters on the registration error. Top row: influence of $r$ with $\sigma=0.01$; Bottom row: influence of $\sigma$ with $r=5$. The vertical bar represents the standard deviation of the error.}
\label{fig:influence_uncertainty}
\end{figure*}

\begin{figure*}[!t]
  \centering
  \begin{tabular}[t]{m{40pt}@{\hspace{0pt}}m{95pt}@{\hspace{0pt}}m{95pt}@{\hspace{0pt}}m{95pt}@{\hspace{0pt}}m{95pt}}
  &\multicolumn{2}{c}{$\sigma=0.01$, $r=10$} & \multicolumn{2}{c}{$\sigma=0.05$, $r=5$}\\
  \centering Our\par method &
  \includegraphics[width=90pt]{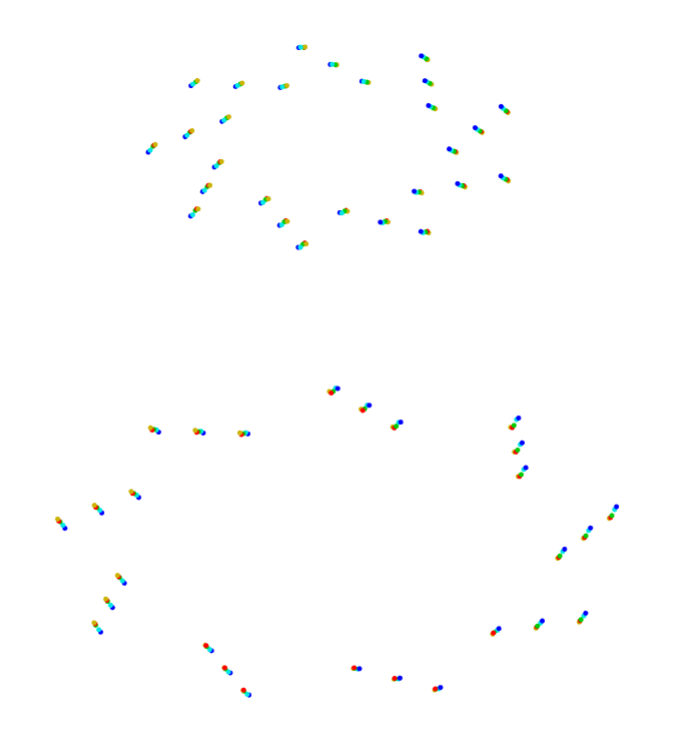} &
  \includegraphics[width=90pt]{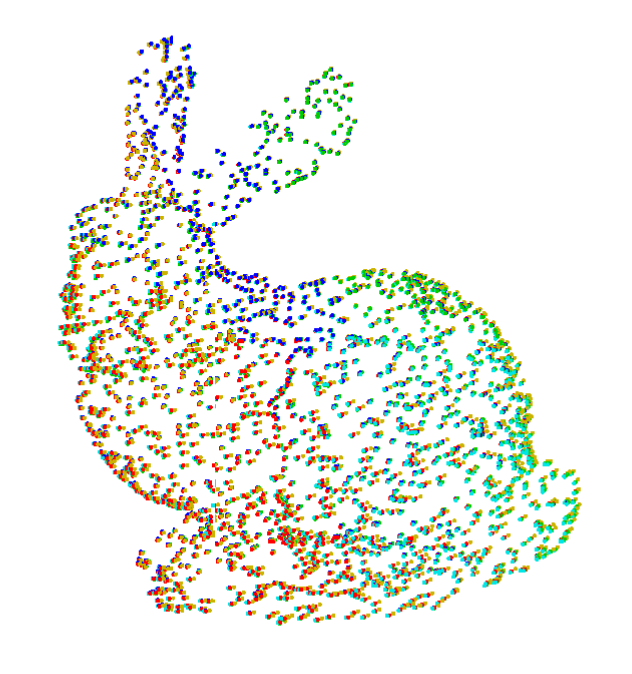} &
  \includegraphics[width=90pt]{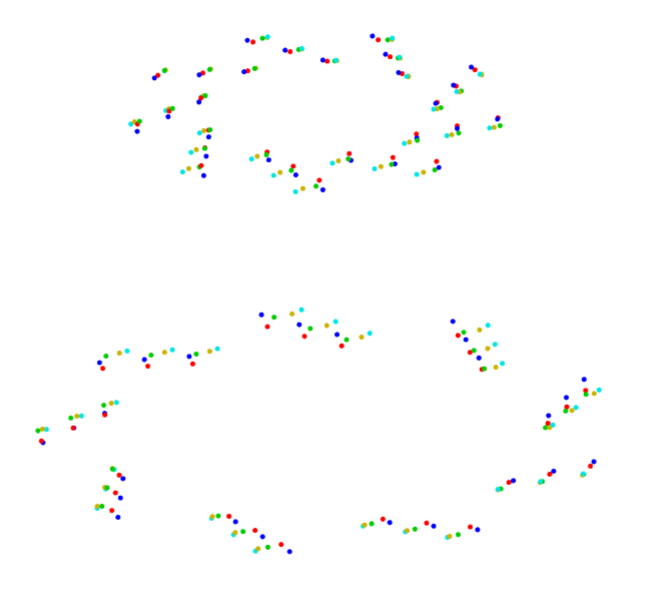} &
  \includegraphics[width=90pt]{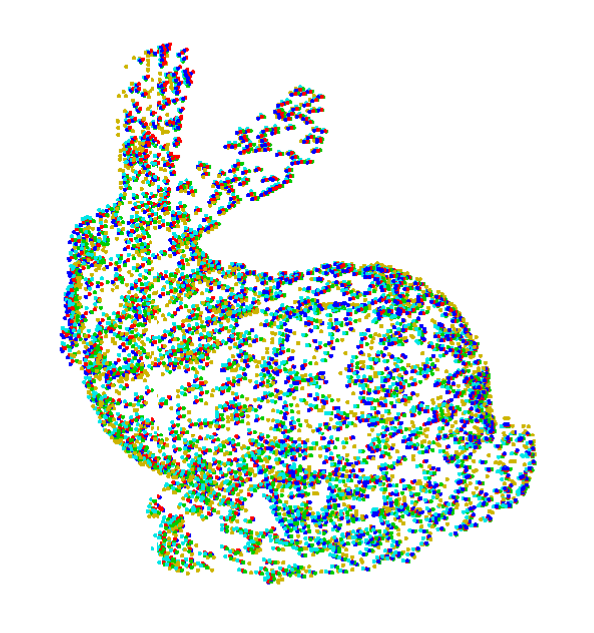} \\
&	\centering\mbox{error = 1.52} & \centering\mbox{error = 0.72} & \centering\mbox{error = 3.30} & ~~~~~~~~\mbox{error = 2.93} \\
  JRMPC&
  \includegraphics[width=90pt]{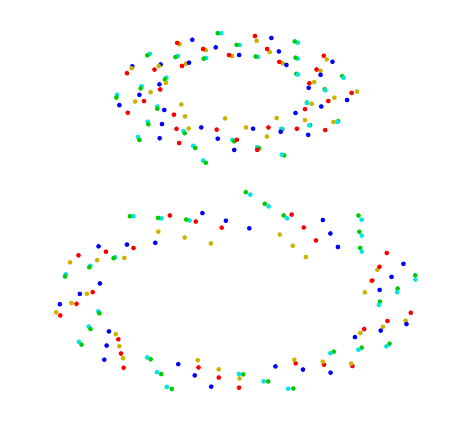} &
  \includegraphics[width=90pt]{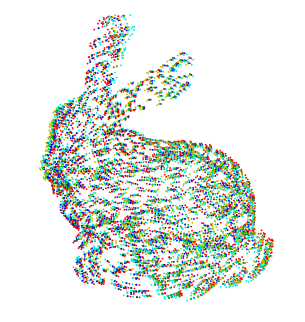} &
  \includegraphics[width=90pt]{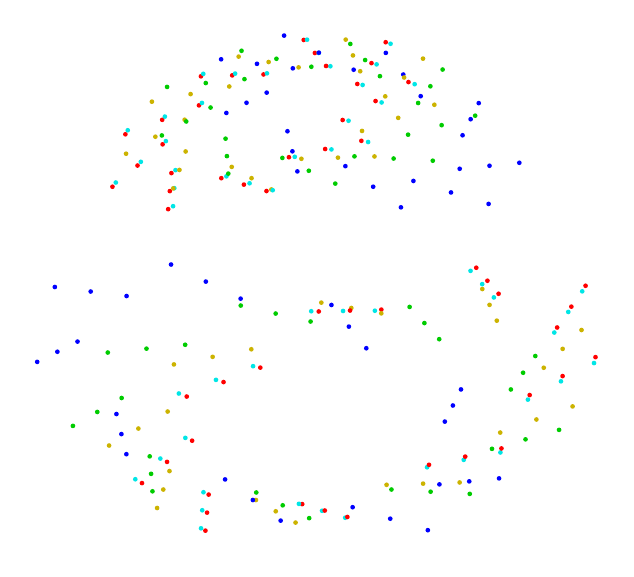} &
  \includegraphics[width=90pt]{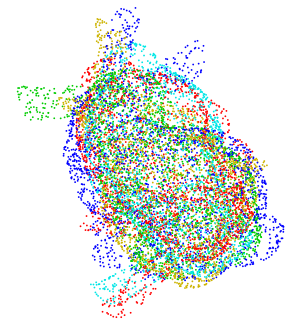} \\
&	\centering\mbox{error = 12.9} & \centering\mbox{error = 4.3} & \centering\mbox{error = 20.1} & ~~~~~~~~\mbox{error = 50.2} 
  \end{tabular}
\caption{Visual results of the registration of 5 point clouds generated from the \textit{triplets} and \textit{bunny} models, for different levels of noise. We show the ground truth models that were used to generate the point clouds (each point cloud has a different color), registered with our method (top row) and JRMPC (bottom row). We consider two combinations of noise parameters: $(\sigma=0.01$, $r=10)$ on the first two columns and $(\sigma=0.05$, $r=5)$ on the last two columns. The quantitative error is given for each case.}
\label{fig:registration}
\end{figure*}

\begin{figure*}[h]
  \centering
  $\begin{array}{m{72pt}@{\hspace{0pt}}m{72pt}@{\hspace{0pt}}m{72pt}@{\hspace{0pt}}m{72pt}@{\hspace{0pt}}m{72pt}@{\hspace{0pt}}m{72pt}}
  \centering\mbox{Ground truth} & \centering\mbox{Our method} & \centering\mbox{JRMPC} &   \centering\mbox{Ground truth} & \centering\mbox{Our method} & ~~~~~\mbox{JRMPC} \\ 
  \includegraphics[width=70pt]{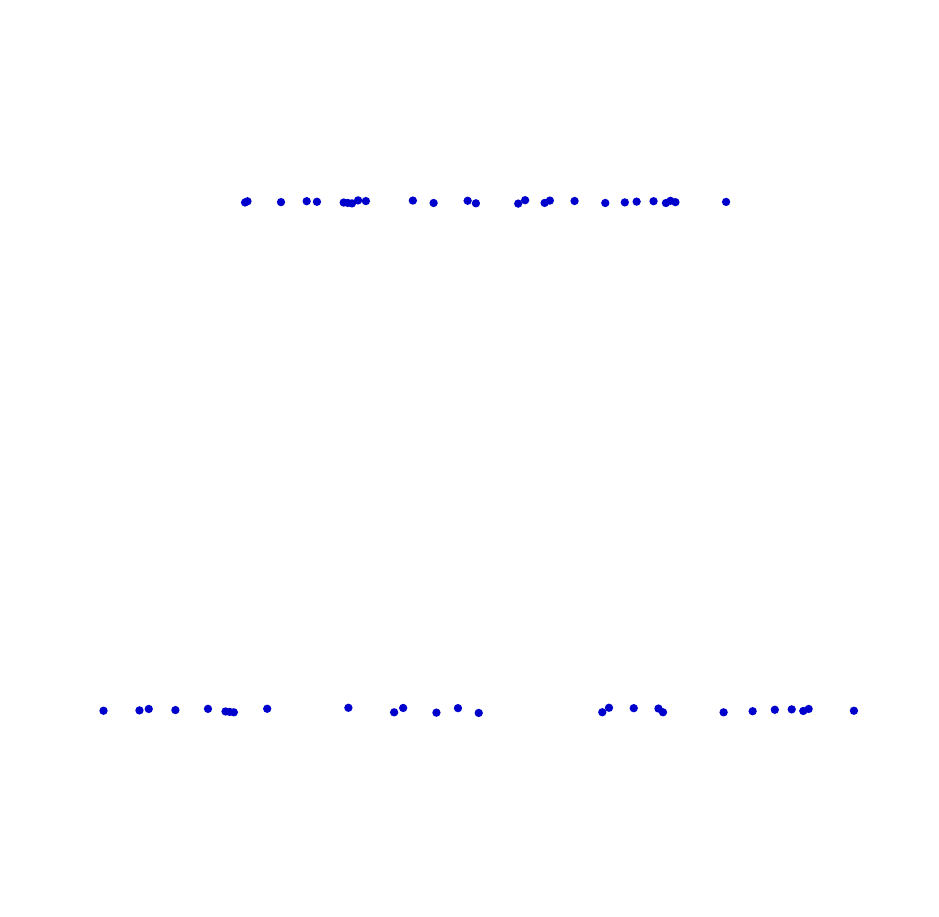} &
  \includegraphics[width=70pt]{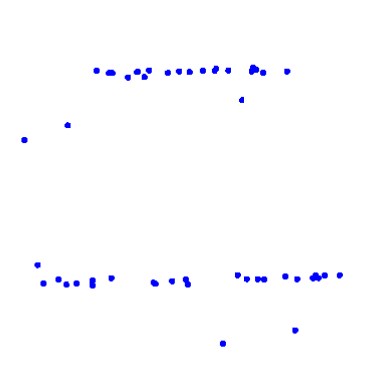} &
  \includegraphics[width=70pt]{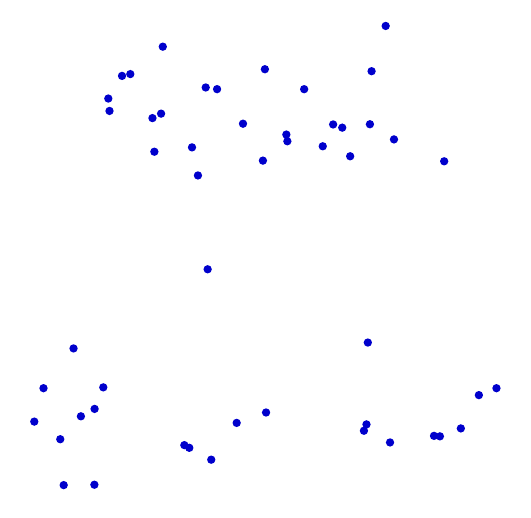} &
  \includegraphics[width=70pt]{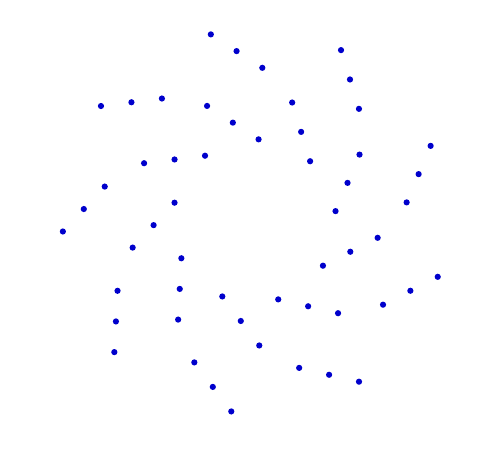} &
  \includegraphics[width=70pt]{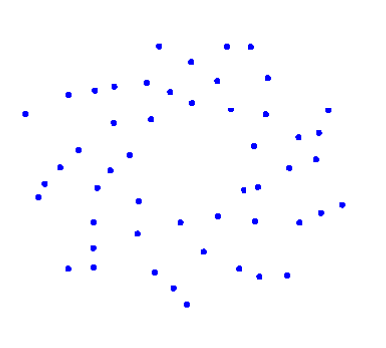} &
  \includegraphics[width=70pt]{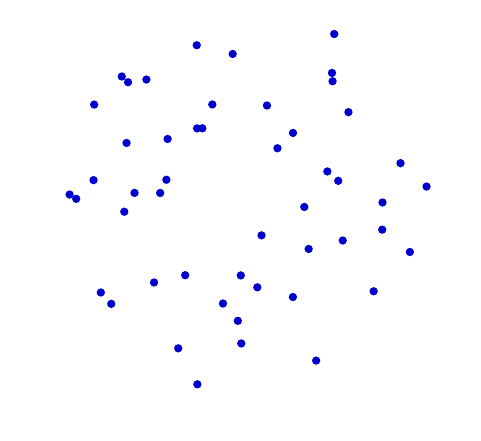} \\
  \includegraphics[width=70pt]{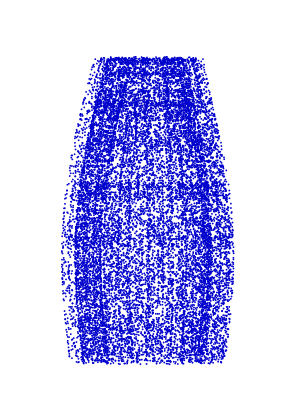} &
  \includegraphics[width=70pt]{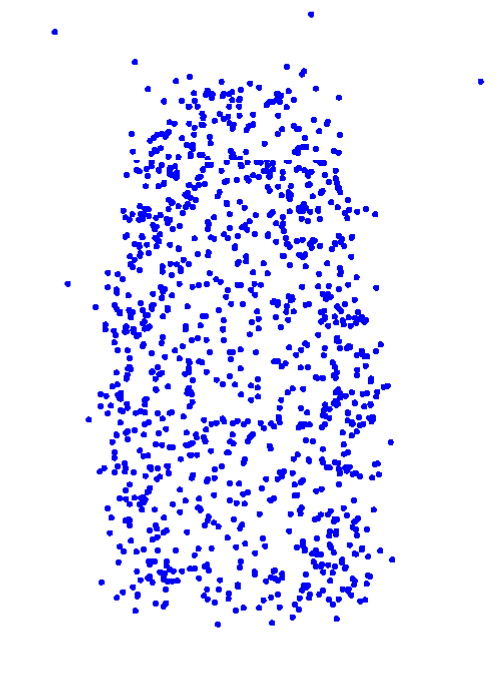} &
  \includegraphics[width=70pt]{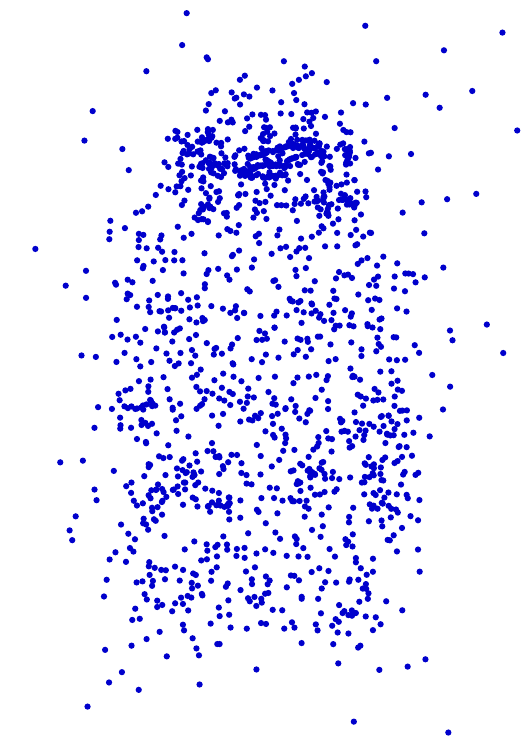} &
  \includegraphics[width=70pt]{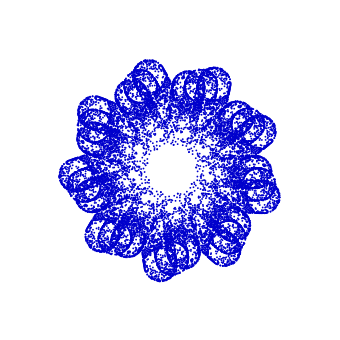} &
  \includegraphics[width=70pt]{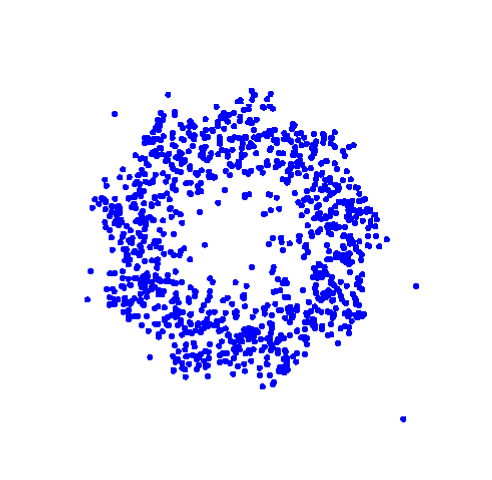} &
  \includegraphics[width=70pt]{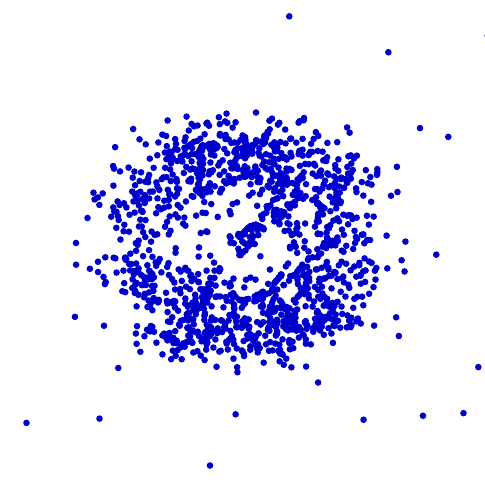} \\
  \includegraphics[width=70pt]{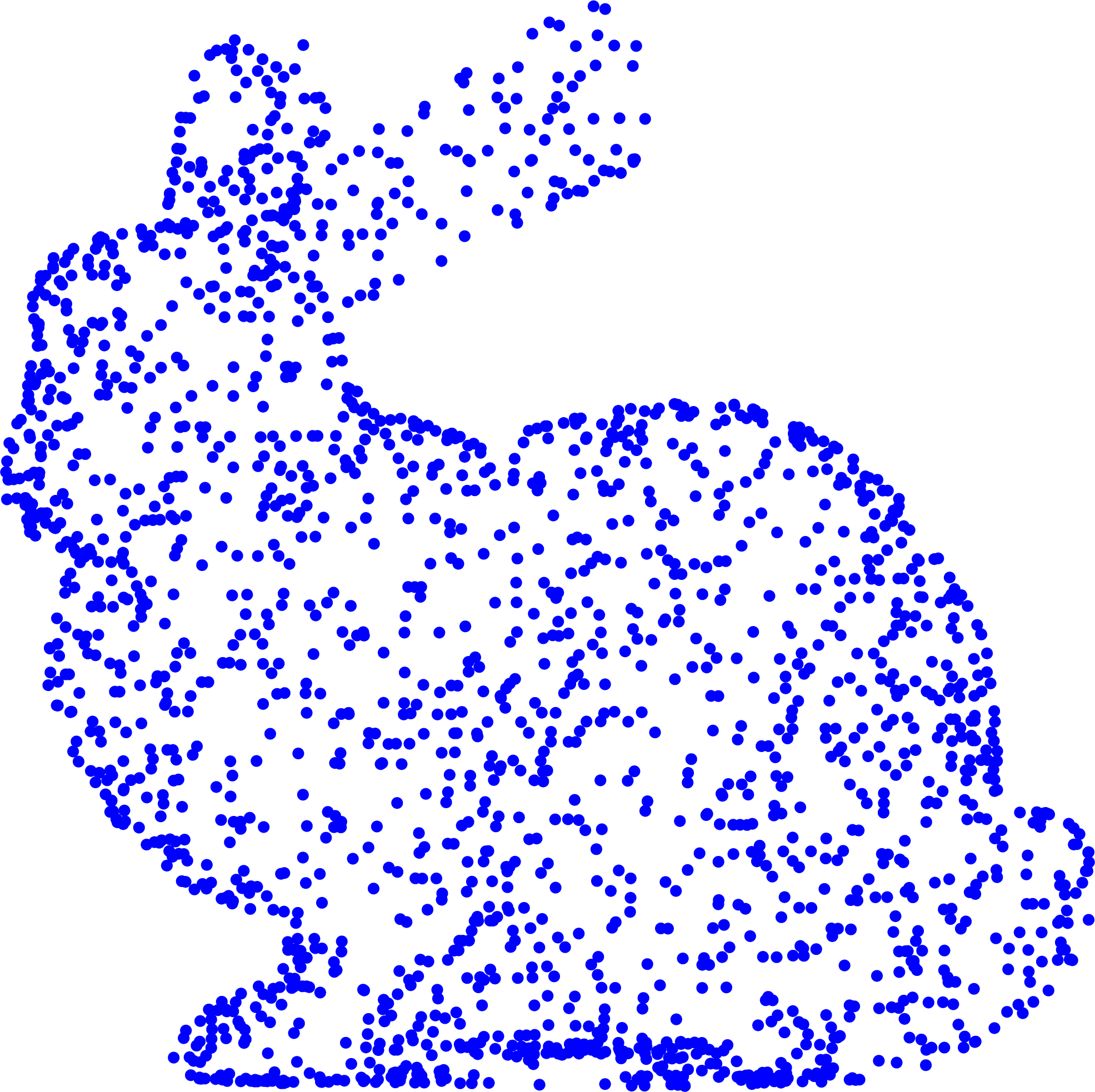} &
  \includegraphics[width=70pt]{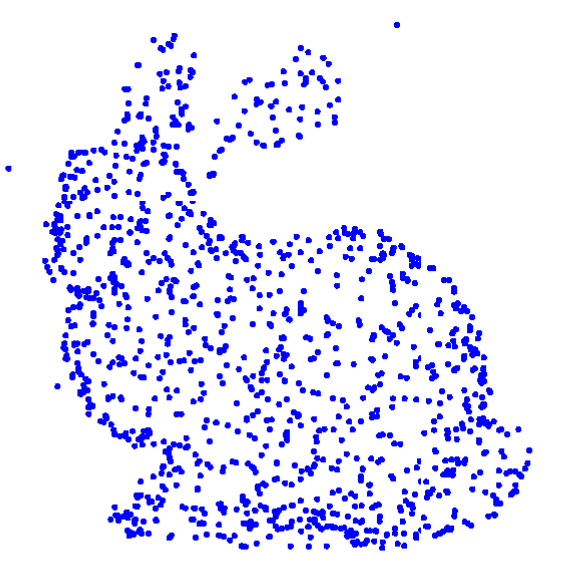} &
  \includegraphics[width=70pt]{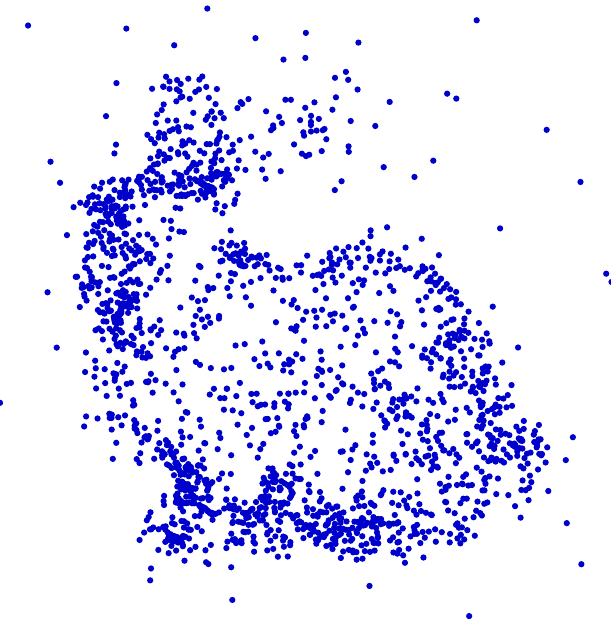} &
  \includegraphics[width=70pt]{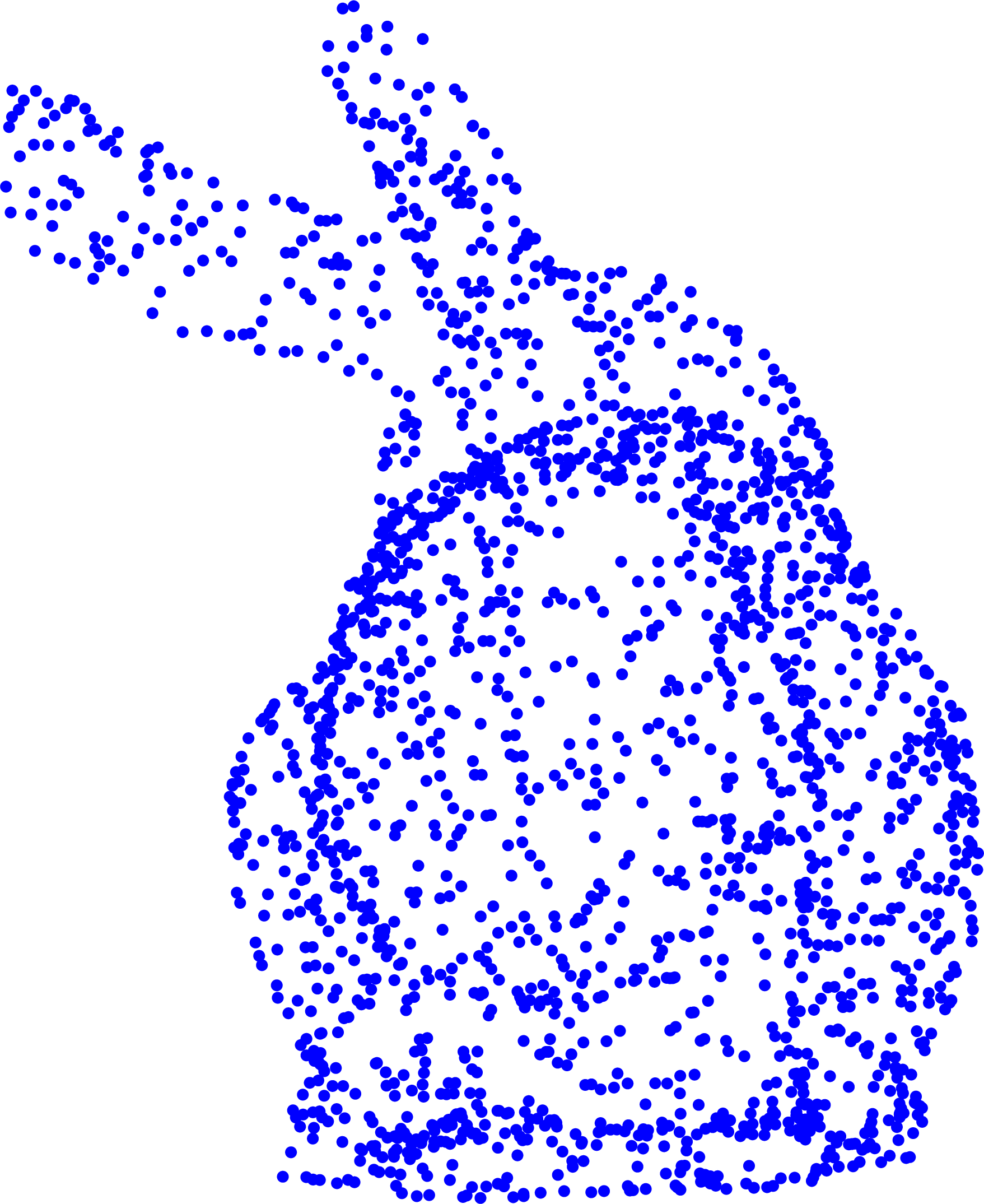} &
  \includegraphics[width=70pt]{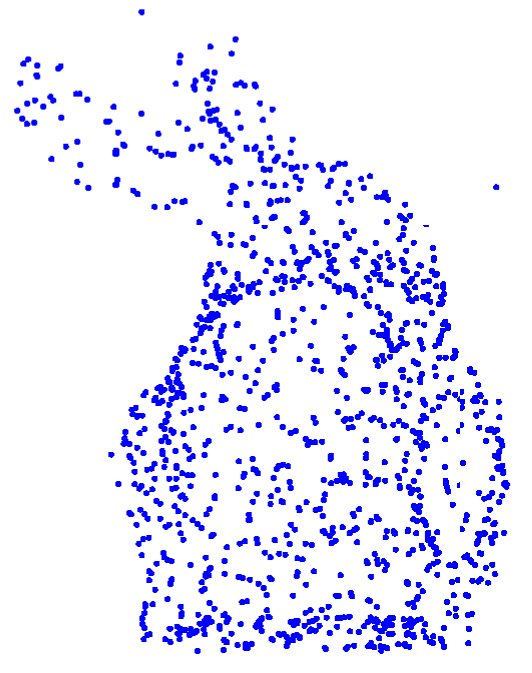} &
  \includegraphics[width=70pt]{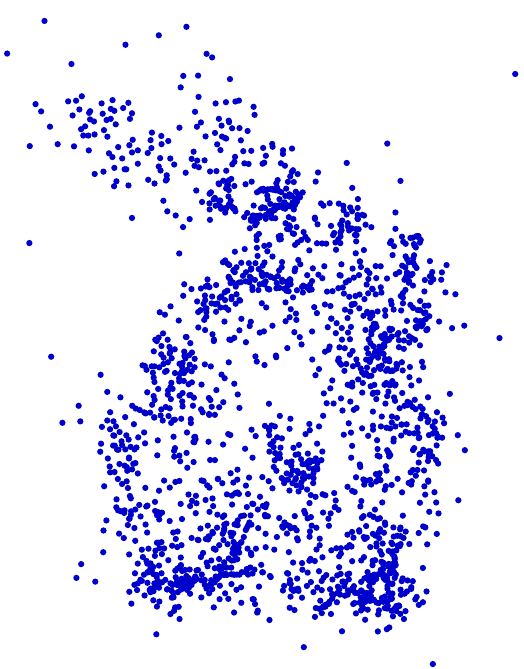} \\
  \end{array}$
\caption{Comparison of the centers of the estimated GMM obtained with our method and JRMPC for the \textit{triplets} (top row), \textit{centriole} (middle row), and \textit{Bunny} (bottom row). Two different views are shown in each case: the first view is on the first three columns and the second view is on the last three columns. The noise parameters have been set to $\sigma=0.01$ and $r=5$.}
\label{fig:reconstruction}
\end{figure*}

\subsubsection{Analysis of the results}

`We are particularly interested in the robustness to noise, which is determined by the noise level $\sigma$ and the anisotropy parameter $r$ defined in \eqref{eq:noise_cov}. In Figure~\ref{fig:influence_uncertainty}, we show the influence of $\sigma$ and $r$ on the registration error for our method and JRMPC. As expected, we first observe that the two methods provide comparable results when the anisotropy parameter $r$ is equal to 1. Indeed, the implicit assumption of JRMPC is isotropic Gaussian noise. The differences for $r=1$ can be explained by the fact that we apply Gaussian noise with spatially varying covariance, which violates the model of JRMPC. We also observe that when the anisotropy parameter $r$ grows (with a fixed  noise level $\sigma$), the error increases much slower with our method than with JRMPC (top line of Figure~\ref{fig:influence_uncertainty}). This highlights the benefit of the proposed modeling. Finally, the same behavior occurs when the noise level $\sigma$ increases with a fixed $r$ (bottom line of Figure~\ref{fig:influence_uncertainty}). For low noise levels, the original data is only slightly corrupted by noise and the influence of the anisotropy parameter is negligible, whereas for higher noise levels the influence of the anisotropy parameter becomes preponderant, which explains the better performances of our method.

In Figure~\ref{fig:registration}, we illustrate the visual quality of the registration results associated to the quantitative errors of Figure~\ref{fig:influence_uncertainty}. We do not show the superimposition of the aligned noisy point clouds, because the high level of noise prevents from a clear visual inspection. To circumvent this problem, we show in Figure~\ref{fig:registration} the aligned ground truth models: the registration has been performed on noisy data, but we show the result applied on ground truth models, in order to be able to visualize the registration errors. Each color corresponds to a different point cloud.
 We consider two combinations of  noise parameters (($\sigma=0.01$, $r=10$) and ($\sigma=0.05$, $r=5$)), and we give the associated quantitative error. In the case ($\sigma=0.01$, $r=10$), the registration with our method is almost perfect, while the results of JRMPC are less accurate. In the higher noise regime ($\sigma=0.05$, $r=5$), our results stay visually satisfying, but JRMPC completely fails to estimate correct registration parameters. 

It is also interesting to look at the GMM density estimated jointly with the registration parameters. We show in Figure~\ref{fig:reconstruction} the centers of the Gaussian components of the GMM estimated with our method and JRMPC, for a medium noise level defined by $\sigma=0.01$ and $r=5$. We can see in each case that the GMM estimated with our method recovers sharper details of the original shapes. 

\begin{table}[!t]
\label{table:K}
    \centering
    \begin{tabular}{|c|c|c|c|c|c|}
        \hline
         $K$ & 500 & 1000 & 1500 & 2000 & 2500\\
        \hline
        Our method & 0.3135 & 0.2934 & 0.2998 & 0.2862 & 0.2813\\
        \hline        
        JRMPC & 0.5455 & 0.6308 & 0.6631 & 0.7064 & 0.7475\\
        \hline        
    \end{tabular}
    \caption{Impact of the number of Gaussian components $K$ on the registration error. The results are obtained on the bunny point cloud, with $\sigma=0.01$, $r=5$ and $M=10$.}
    \label{tab:my_label}
\end{table}

These results show how the integration of our noise model in the EM estimation improves both numerically and visually the registration results. This is explained by the fact that with JRMPC, each Gaussian component of the GMM has to account both for the shape of the object and the noise. Therefore, the reconstructed GMM of Figure~\ref{fig:reconstruction} tends to fit the noise with JRMPC. Our method decouples noise handling from shape modeling, such that the GMM actually fits only the shape of the objects in Figure~\ref{fig:reconstruction}. Since the GMM acts as a target for the registration of the input point clouds in the EM algorithm, the quality of the registration is directly intertwined with the quality of the GMM reconstruction.

In Table \ref{table:K}, we show the impact of the number of Gaussian components $K$ on the registration error. The accuracy of our method depends only marginally on the value of $K$. In contrast, the error of JRMPC slightly increases with the number of Gaussians. This could be explained as follows: since the GMM distribution estimated by JRMPC fits the noise of the data, increasing the number of Gaussians improves this noise fitting and consequently degrades the registration. On the other hand, our method is more computationally expensive with a runtime of approximately 9 minutes, whereas the runtime of JRMPC is about 1 minute, for $K=500$ on the {\it bunny} point cloud. %However, it is necessary to consider a high value of $K$ to obtain satisfying reconstruction (see Figure \ref{fig:reconstruction}). }

%\todo{Autres expérimentations possibles: experiments with space varying isotropic noise, gmm fitting with known rotations, comparison with other (pairwise) methods}	

%\todo{Computation time?}	

%\begin{figure*}[t]
%  \centering
%  $\begin{array}{c@{\hspace{0pt}}c@{\hspace{0pt}}c}
%  \includegraphics[width=175pt]{figures/influence_M_triplets.pdf} &
%  \includegraphics[width=175pt]{figures/influence_M_centriole.pdf} & 
%  \includegraphics[width=175pt]{figures/influence_M_bunny.pdf} \\ 
%  \includegraphics[width=175pt]{figures/influence_o_triplets.pdf} &
%  \includegraphics[width=175pt]{figures/influence_o_centriole.pdf} & 
%  \includegraphics[width=175pt]{figures/influence_o_bunny.pdf} \\ 
%  \mbox{\it{Triplets}} & \mbox{\it{Centriole}} & \mbox{\it{Bunny}} \\ 
%  \end{array}$
%\caption{Influence of the level of uncertainty. To row: influence of $r$; Bottom row: influence of $\sigma$. The vertical bar represents the variance of the error.}
%\label{fig:influence_uncertainty}
%\end{figure*}
%

\subsection{Real data}

\begin{figure}[t]
  \centering
  $\begin{array}{m{120pt}@{\hspace{0pt}}m{120pt}}
  \includegraphics[width=120pt]{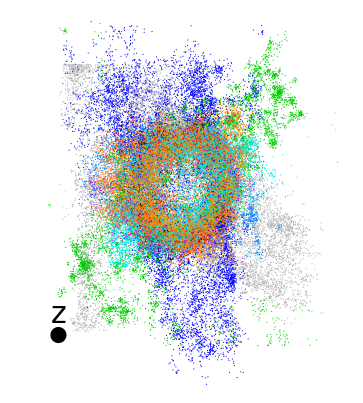} &
  \includegraphics[width=120pt]{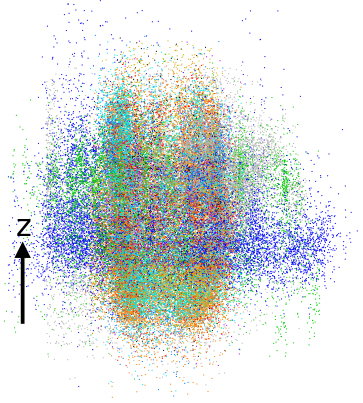} 
  \end{array}$
\caption{Superimposition of the ten real dStorm acquisitions of centriole in their original pose (the red and blue point clouds are presented individually in Figure~\ref{fig:real_data}). The first column represents a view orthogonal to the $z$ direction, and the second column represents a view parallel to the $z$ direction.}
\label{fig:real_data_all}
\end{figure}

\begin{figure*}[!t]
  \centering
  \begin{tabular}{m{50pt}@{\hspace{0pt}}m{120pt}@{\hspace{0pt}}m{140pt}@{\hspace{0pt}}m{85pt}}
  \centering Our \par method&
  \includegraphics[width=120pt]{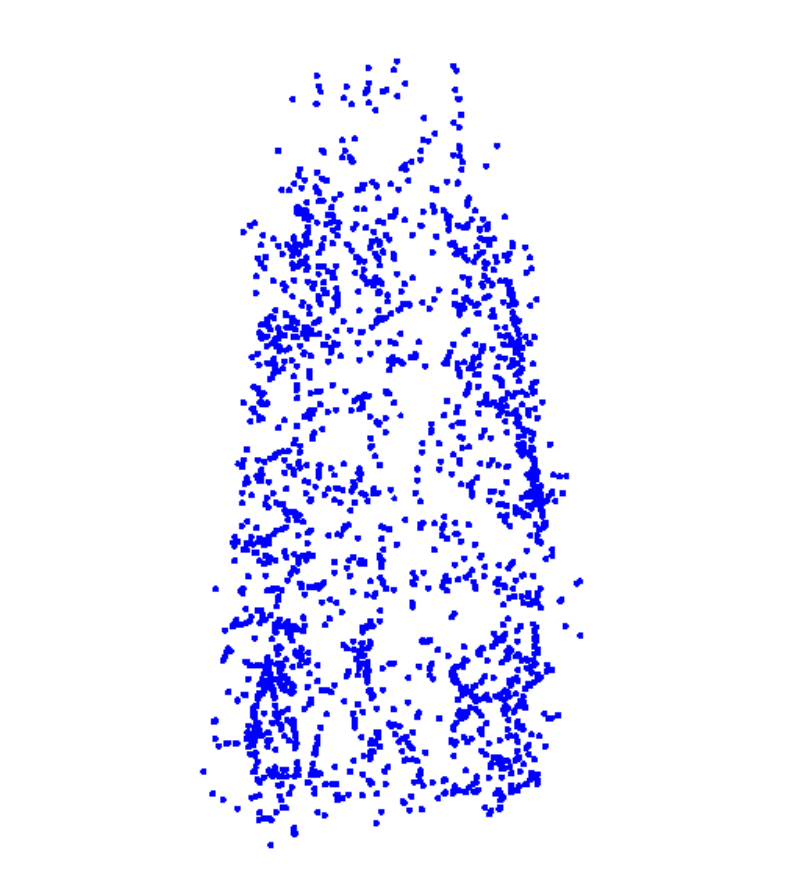} &
  \includegraphics[width=120pt]{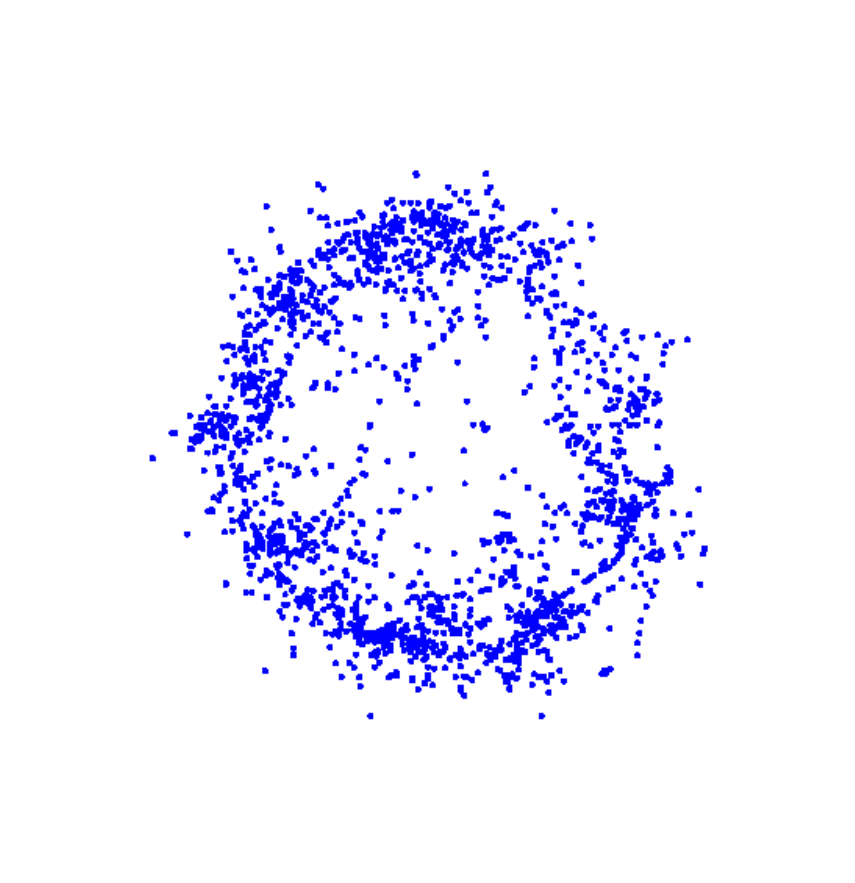} &
  \includegraphics[width=80pt]{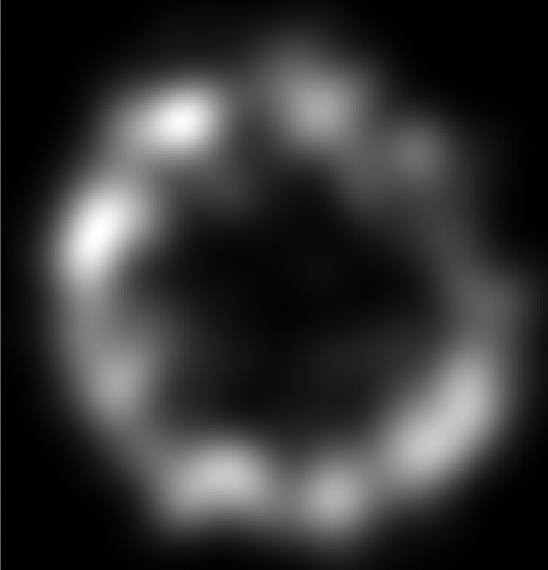} \\
  &\centering\mbox{(a)} & \centering\mbox{(b)}& ~~~~~~~~~~~~~~\mbox{(c)} \\
  JRMPC&
  \includegraphics[width=120pt]{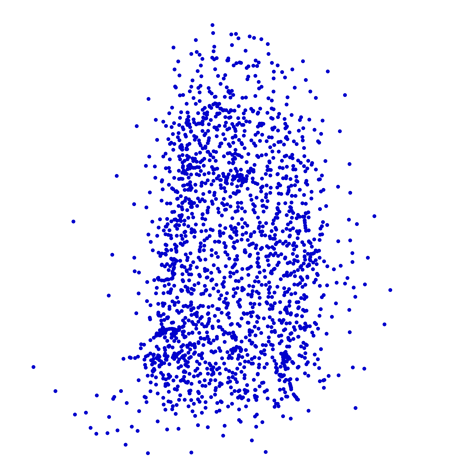} &
  \includegraphics[width=120pt]{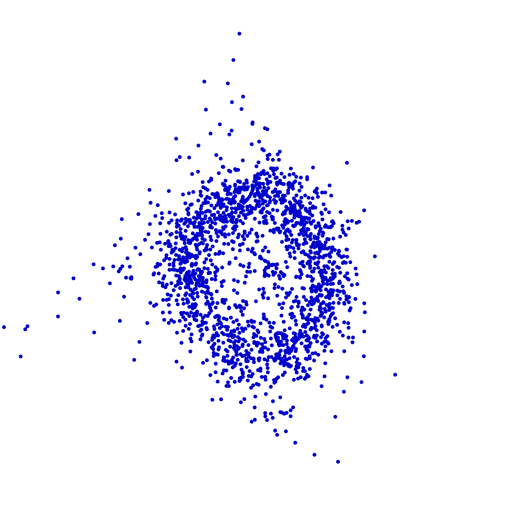} &
  \includegraphics[width=80pt]{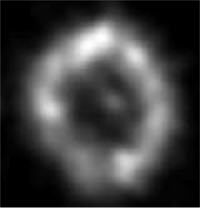} \\
  &\centering\mbox{(d)} & \centering\mbox{(e)}& ~~~~~~~~~~~~~~\mbox{(f)} \\[10pt]
  \end{tabular}
\caption{Centers of the estimated GMM on real data. The first two columns are two orthogonal views. The last column shows the projection in the direction of the centriole axis of the volume obtained by evaluating the GMM on a 3D grid.}
\label{fig:real_gmm}
\end{figure*}

We applied our method on real microscopy data. We used direct stochastic optical reconstruction microscopy (dStorm), which one of the most popular SMLM techniques, combined with expansion microscopy \cite{Zwettler20}. We insist on the fact that the noise covariances $\bSigmaji$ associated to each point of the dStorm acquistion are known. Each point  corresponds to the localization of a fluorophore, obtained with a detection algorithm applied on images acquired with a conventional microscope (for details about SMLM acquisition and detection algorithms, see \cite{Sage19}). The covariances $\overline{\boldsymbol{\Sigma}}_{ji}$ are the localization uncertainties provided by the detection algorithm. 
It produces point clouds with a known anisotropic localization uncertainty of the form (\ref{eq:noise_cov}). We imaged the tubulin structure of the centriole, which is a macromolecular assembly that plays a crucial role in several cellular mechanisms and has motivated a lot of recent research to decipher its structural properties \cite{LeGuennec21}. It is know to have a barrel-like shape and a C9 symmetry structure, from which our \textit{centriole} simulated structure is inspired from (see Figure~\ref{fig:simulated_data}). The acquired point cloud contained several centrioles, and we manually cropped $M=10$ of them, shown in Figure~\ref{fig:real_data}.e-f, to perform registration. The number of point in each point cloud varies between 2013 and 11166, and we use $K=2000$ for the GMM reconstruction. As already discussed in Section \ref{sec:intro} and Figure~\ref{fig:real_data},  we can see that the noise in the axial direction z of the microscope is much higher than in the lateral plane. The localization noise of each point is known and follows a Gaussian law with a covariance of the form (\ref{eq:noise_cov}). The experimental distribution of the anisotropy parameter $r$ can be approximated by a Gaussian law with mean 5 and standard deviation 1.5. Therefore, it is crucial to combine the information of each view to reconstruct high resolution particles. Fig \ref{fig:real_data_all} shows all the input point clouds in their original pose. We initialize the registration parameters with the coarse registration method dedicated to centrioles described in \cite{Fan21}.  

The reconstruction of the GMM is of particular interest in our application since our goal is to reconstruct a model of the particle with a higher resolution than in the original data. The estimated centers of the GMM can serve as such a reconstruction. In Figure~\ref{fig:real_gmm}, we show the centers estimated with our method and JRMPC. Similarly to the simulated case, we observe that our method captures more precisely the shape of the centriole, and the reconstruction of JRMPC is more influenced by the presence of noise. To obtain Figure~\ref{fig:real_gmm}.c and \ref{fig:real_gmm}.f, we created volumes by summing Gaussians of fixed variance centered on each of the estimated centers, and we projected the volumes in the direction of the symmetry axis of the centriole. With our method we can recover the ninefold symmetry of the centriole in Figure~\ref{fig:real_gmm}.c, whereas the JRMPC result is much more blurred. This shows that our noise handling approach can reveal biologically relevant information that is lost with the baseline method.

\begin{figure*}[!ht]
  \centering
  \begin{tabular}[t]{m{40pt}@{\hspace{0pt}}m{96pt}@{\hspace{0pt}}m{96pt}@{\hspace{0pt}}m{96pt}@{\hspace{0pt}}m{96pt}}
  &\multicolumn{2}{c}{Original data} & \multicolumn{2}{c}{Cleaned data}\\
  \centering Our\par method& 
  \includegraphics[width=96pt]{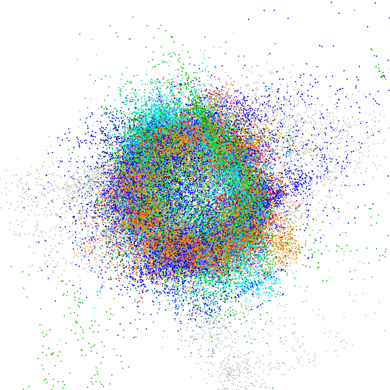} &
  \includegraphics[width=96pt]{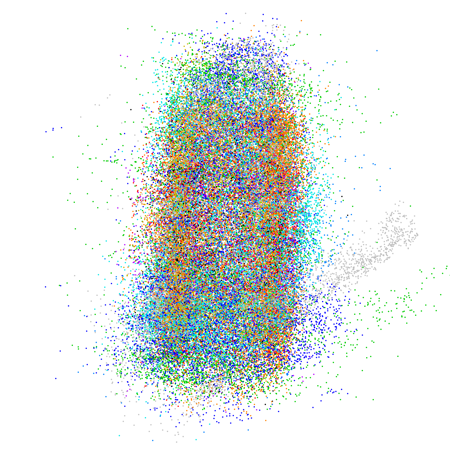} &
  \includegraphics[width=96pt]{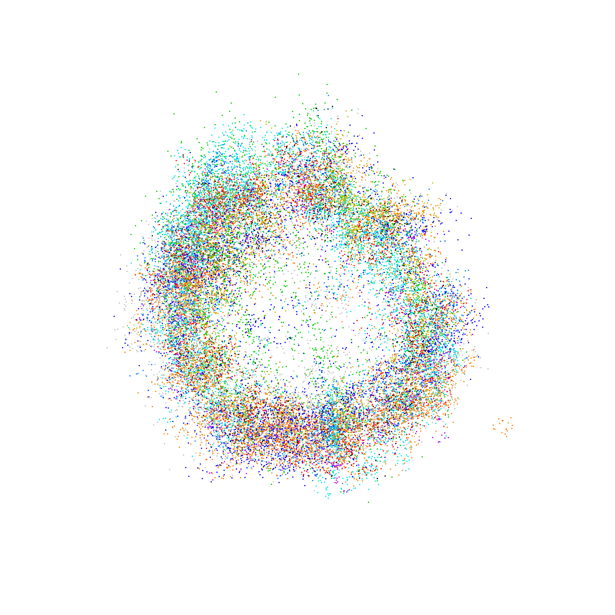} &
  \includegraphics[width=96pt]{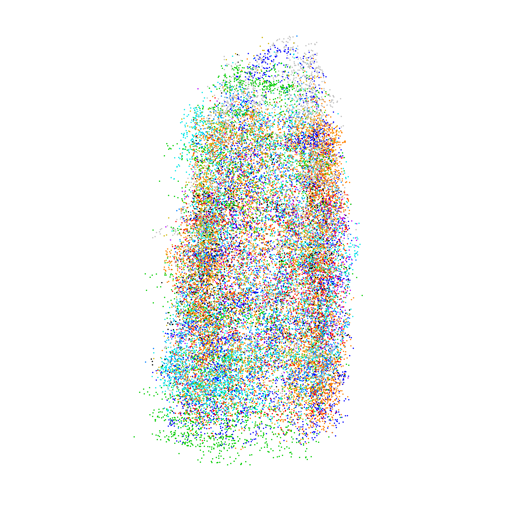} \\
  &\centering\mbox{(a)} & \centering\mbox{(b)}& \centering\mbox{(c)}& ~~~~~~~~~~~~~~~\mbox{(d)}\\
  JRMPC&
  \includegraphics[width=96pt]{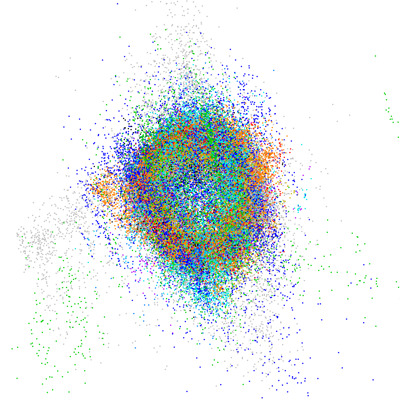} &
  \includegraphics[width=96pt]{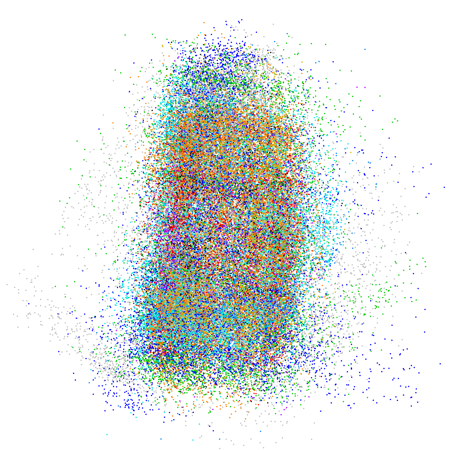} & 
  \includegraphics[width=96pt]{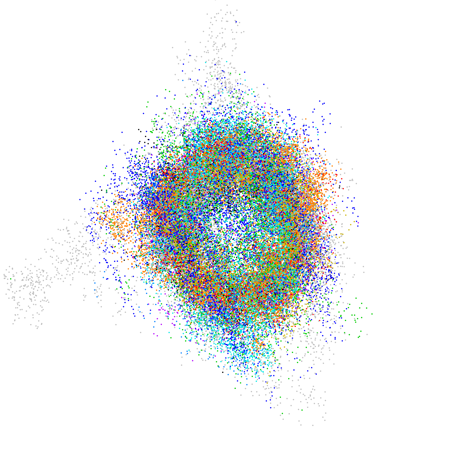} &
  \includegraphics[width=96pt]{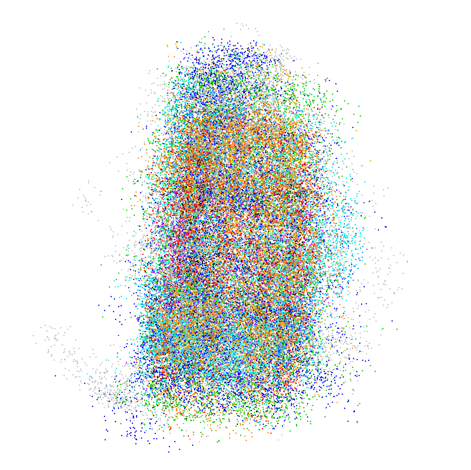} \\
  &\centering\mbox{(e)} & \centering\mbox{(f)}& \centering\mbox{(g)}& ~~~~~~~~~~~~~~~\mbox{(h)} \\[10pt]
  \end{tabular}
\caption{Registered point clouds from the real data of Figure~\ref{fig:real_data}. In the first two columns, we show the original point clouds, and in the last two columns, we show the cleaned point clouds obtained after post processing.}
\label{fig:real_registration}
\end{figure*}

In Figure~\ref{fig:real_registration}.a,b,e,f we show the registered point clouds obtained with the two methods. It is difficult to visually assess the quality of the registration because of the high amount of noise. Therefore, we show in In Figure~\ref{fig:real_registration}.c,d,g,h cleaned versions of the registered point clouds. They are obtained by applying a postprocessing %described in \cite{Evangelidis17} 
that removes points classified as outliers and points assigned to Gaussian components with high variance (superior to 2.5 times the median variance). A Gaussian component with high variance means that it does not cluster accurately the shape of the object and is likely to also capture noise. We see that the cleaned point clouds remain very noisy with JRMPC, while our clean data recovers the shape of the centriole. This means that the variances of the GMM clusters do not capture noise and are concentrated on the particle.

\section{Conclusion}
We presented a novel method to handle anisotropic and spatially varying noise in the EM-GMM framework for point clouds registration. Unlike previous methods, we define an explicit Gaussian prior on the localization noise. We derived closed form updates for arbitrary anisotropic covariances at each step of our EM algorithm. Experimental results showed that our method outperforms the baseline approach \cite{Evangelidis17} when the point clouds are corrupted with high anisotropic noise, both in terms of registration accuracy and GMM reconstruction. We also emphasize the practical interest of our method on real SMLM data, for which there is an urgent need of registration methods taking into account high localization uncertainty.

\newpage
\appendix
\section{Computation of $p({\byji}|\theta)$}
\label{sec::app1}

The term $$p({\byji}|\theta)  = \int_{\yji} p(\byji|\yji) p(\yji|\theta)d\yji$$
can be written
\begin{eqnarray}
\label{eq:likelihood_noise_def}
p({\byji}|\theta)  &=& \sum_{k=1}^K p_k \int_{\yji} p(\byji|\yji)  p(\yji|\zji=k, \theta)d\yji \\
    &=& \sum_{k=1}^K p_k \int_{\yji}  \mN(\byji;\yji,\bSigmaji) \mN(\phi_j(\yji);\bmu_k,\bSigma_k) d\yji 
\end{eqnarray}

Using 
\begin{eqnarray*}
\mN(\byji;\yji,\bSigmaji) &=& \mN(\phi_j(\byji);\phi_j(\yji),\bSigmaji^{\R_j})\\ 
    &=& \mN(\phi_j(\byji)-\phi_j({\yji});0,\bSigmaji^{\R_j}),
\end{eqnarray*}
where $\bSigmaji^{\R_j}=\R_j\overline\bSigma_{ji}\R_j^T$, and with the change of variables ${\y}'_{ji} = \phi_j(\yji)$, we have:
\begin{equation}
\begin{split}
p({\byji}|\theta) = \sum_{k=1}^K p_k \int_{\y'_{ji}}  \mN(\phi_j(\byji)-\y'_{ji};0,\bSigmaji^{\R_j})  \, \mN(\y'_{ji};\bmu_k,\bSigma_k) d\y'_{ji} \enspace.
\end{split}
\end{equation}
The expression of $p({\byji}|\theta)$ in (\ref{eq:L2}) can be obtained easily using properties related to the sum of two normally distributed random variables.

\section{Derivation of $Q(\theta,\theta^{(l)})$}
\label{sec:app2}
From the definition of the complete-data likelihood  (\ref{eq:L4}), the auxiliary function defined in (\ref{eq:Qdef}) writes
\begin{equation}
\label{eq:tm1}
\begin{split}
Q(\theta,\theta^{(l)}) = \sum_\z\int_{\mathbb \y} p(\y,\z|\by,\theta^{(l)}) \log(p(\bar\y|\y) )d\y + \sum_\z\int_{\mathbb \y} p(\y,\z|\by,\theta^{(l)}) \log(p(\y,\z|\theta)d\y.
\end{split}
\end{equation}
The first term in (\ref{eq:tm1}) does not depend on $\theta$ and the second one can be factorized using the same strategy than 
for the standard the GMM case \cite{Bilmes98}. 
We obtain
\begin{equation}
\begin{split}
Q(\theta,\theta^{(l)}) = (constant) + \sum_{jik}\Big[\int_{\yji} p(\yji,\zji=k|\byji,\theta^{(l)})  \log(p(\yji,\zji=k|\theta) d{{\yji}} \Bigg].
\end{split}
\end{equation}

Since $\alphajik \beta_{jik}  (\yji )=  p(\yji,\zji=k|\byji,\theta^{(l)})$ (see \eqref{eq:ab}), we get 
\begin{equation}
\begin{split}
Q(\theta,\theta^{(l)})= (constant) + \sum_{jik} \alphajik \Bigg[\int_{\yji} \beta_{jik}  (\yji )  \log(p(\yji,\zji=k|\theta) d{{\yji}} \Bigg]\enspace.
\label{eq:Qdemo}
\end{split}
\end{equation}

To obtain the result \eqref{eq:Q}, we use 
\begin{eqnarray*}
p(\yji,\zji=k|\theta) &=& p_k \,p(\yji|\zji=k,\theta) \\
    &=& p_k \,\mN(\phi_j(\yji);\bmu_k,\bSigma_k)\enspace.
\end{eqnarray*}

\section{Computation of $\beta_{jik}$}
\label{sec:app3}
Using Bayes theorem, we have
\begin{equation*}
\begin{split}
\beta_{jik}(\yji) &\propto p(\byji|\yji) p(\yji|\zji=k,\theta^{(l)})   \\
&\propto \mN(\byji;\yji,\bSigmaji) \mN(\phi_j(\yji);\bmu_k,\bSigma_k) \\
\end{split}
\end{equation*}
Since $$\mN(\byji;\yji,\bSigmaji) = \mN(\phi_j(\byji);\phi_j(\yji),\bSigmaji^{\R_j})$$ 
and  $$\mN(a;b,c)=\mN(b+d;a+d,c)\enspace,$$ we 
can obtain easily
\begin{equation}
\begin{split}
\beta_{jik}(\yji) \propto 
\mN(\phi_j(\yji)-\bmu_k;\phi_j(\byji)-\bmu_k,\bSigmaji^{\R_j})\,
\mN(\phi_j(\yji)-\bmu_k;0,\bSigma_k).
\end{split}
\label{eq:ll}
\end{equation}

We use the property $$\mN(\x;\bmu_1,\bSigma_1)\mN(\x;0,\bSigma_2)\propto\mN(.;\bmu_3,\bSigma_3)$$ with  $\bmu_3=\bSigma_2(\bSigma_1+\bSigma_2)^{-1}\bmu_1$, and $\bSigma_3=\bSigma_1(\bSigma_1+\bSigma_2)^{-1}\bSigma_2$, to obtain
\begin{equation}
\beta_{jik}(\yji)=\mN(\phi_j(\yji)-\bmu_k;\hat{\y}_{ji}-\bmu_k,  \bSigmaji^{\R_j} \left(\bSigma_k+\bSigmaji^{\R_j}\right)^{-1} \bSigma_k)
\label{eq:mm}
\end{equation}
%Moreover, $\beta_{jik}(\yji)$ is also equal to the term of Eq. \ref{eq:mm} since both terms integrate to 1.

Finally, in order to obtain the  result (\ref{eq:beta}), we have to show that $$(\I-\W_{jik})= \bSigmaji^{\R_j} \left(\bSigma_k+\bSigmaji^{\R_j}\right)^{-1}\enspace.$$ This can be obtained by writing $\I$ as  $$\left(\bSigma_k+\bSigmaji^{\R_j}\right)
\left(\bSigma_k+\bSigmaji^{\R_j}\right)^{-1}\enspace.$$

\section{Update of the GMM parameters}
\label{sec:app4}
The update of the GMM parameters can be  obtained by cancelling the derivative of $Q(\theta,\theta^{(l)}$) in (\ref{eq:Q}).\\

\noindent{\bf Update of $p_k$}~
For the update of $p_k$, we also need to impose the constraint $\sum_k p_k=1$. To this end, we introduce the Lagrange multiplier $\lambda$ and solve the equation:
\begin{equation}
\begin{split}
\frac{\partial}{\partial p_k} \Bigg[
\sum_{ji}  \Bigg(  \alphajik \int_{\yji} 
\beta_{jik}  (\yji )  \log(p_k) d\yji\Bigg)  +  \lambda \Bigg(\sum_{l=1}^K p_l-1\Bigg) \Bigg]=0
\end{split}
\end{equation}
Since $\int_{\yji}\beta_{jik}(\yji) d\yji =1$, we obtain directly the  result (\ref{eq:pk}).\\

\noindent{\bf Update of $\bmu_k$}~
The cancellation of the derivative with respect to $\bmu_k$ yields 
\begin{equation}
\sum_{ji} \alpha_{jik}  \int_{\yji} \beta_{jik}(\yji) \bSigma_k^{-1}(\phi_j(\yji)-\bmu_k)d\yji=0
\label{eq:mmm}
\end{equation}
Since $\int_{\yji}\beta_{jik}(\yji) d\yji =1$, (\ref{eq:mmm}) writes
\begin{equation}
\sum_{ji} \alpha_{jik} \bSigma_k^{-1}\left(\int_{\yji} \phi_j(\yji) \beta_{jik}(\yji)d\yji - \bmu_k \right) =0.
\end{equation}
Using the  notation $E[X] = \int_{y} \beta_{jik}(y) X dy$, we get
\begin{equation}\sum_{ji} \alpha_{jik} \bSigma_k^{-1} \left( E[\phi_j(\yji)] - \bmu_k \right)=0.
\end{equation}
which  can be solved to obtain the expected result (\ref{eq:muupdate}).\\

\noindent{\bf Update of $\bSigma_k$}~
We rewrite (\ref{eq:Q}) without the constant terms w.r.t. $\bSigma_k$ (and follow the strategy desribed in \cite{Bilmes98} for the standard GMM case) to obtain
\begin{equation}
\begin{split}
Q(\theta,\theta^{(l)})&=-\frac{1}{2}\sum_k \Bigg(\log\left|\bSigma_k\right| \sum_{ji} \alpha_{jik} +  \\  
&\sum_{ji} \alpha_{jik} \trace \bigg(  \bSigma_k^{-1}  
\int_{\yji} \beta_{jik}(\yji) (\phi_j(\yji)-\bmu_k) (\phi_j(\yji)-\bmu_k)^T d\yji
\bigg)                                       \Bigg)
\label{eq:appC}
\end{split}
\end{equation}
Using the  notation $E[X] = \int_{y} \beta_{jik}(y) X dy$, and following \cite{Bilmes98}, we can easily obtain the result (\ref{eq:sigmaupdate}).\\

\noindent{\bf Derivation of Algorithm \ref{alg2}}
We first need to compute $E[\phi_j(\yji)]$ and\\ $E[(\phi_j(\yji)-\bmu_k) (\phi_j(\yji)-\bmu_k)^T]$. Based on (\ref{eq:beta}), $E[\phi_j(\yji)]$ can be written
\begin{equation}
\int_{\yji} \phi_j(\yji)  \mN(\phi_j(\yji));\hat{\y}_{ji},(\I-\W_{jik}) \bSigma_k)) d\yji
\end{equation}
Using the change of variable $\y'_{ji} = \phi_j(\yji)$, we get:
\begin{equation}
\begin{array}{lll}
E[\phi_j(\yji)] &=& \int_{\y'_{ji}} \y'_{ji}  \mN(\y'_{ji};\hat{\y}_{ji},(\I-\W_{jik}) \bSigma_k)) d\y'_{ji} \\
 &=& \hat{\y}_{ji}.
 \end{array}
\end{equation}
We can compute $E[\phi_j(\yji) \phi_j(\yji)^T]$ with the same change of variables to obtain
\begin{equation}
\int_{\y'_{ji}} \y'_{ji} {\y'_{ji}}^T 
\mN(\y'_{ji}|\hat{\y}_{ji},(\I-\W_{jik}) \bSigma_k)) d\y'_{ji} =\hat{\y}_{ji} \hat{\y}_{ji}^T + (\I-\W_{jik}) \bSigma_k
\end{equation}
Then, we can then easily derive:
\begin{equation}
\begin{split}
E[(\phi_j(\yji)-\bmu_k) (\phi_j(\yji)-\bmu_k)^T] = 
\hat{\y}_{ji} \hat{\y}_{ji}^T  
+ (\I-\W_{jik}) \bSigma_k 
- \bmu_k \hat{\y}_{ji}^T- \hat{\y}_{ji}^T \bmu_k + \bmu_k \bmu_k^T
\end{split}
\label{eq:plug}
\end{equation}

Plugging (\ref{eq:plug}) into (\ref{eq:sigmaupdate}) and using the updated form of $\bmu_k$ (\ref{eq:muupdate}), we obtain
\begin{equation}
\label{eq:bou}
\bSigma_k=\frac{\sum_{j=1}^M\sum_{i=1}^{N_j}\alphajik 
\left(\hat{\y}_{ji} \hat{\y}_{ji}^T + (\I-\W_{jik}) \bSigma_k \right)} {\sum_{j=1}^M\sum_{i=1}^{N_j}\alphajik} - \bmu_k \bmu_k^T.
\end{equation}

We can then derive Alg. \ref{alg2} from (\ref{eq:pk}) and (\ref{eq:muupdate}) with 
$E[\phi_j(\yji)] = \hat{\y}_{ji}$, and from (\ref{eq:bou}).

\section{Update of the transformation parameters}
\label{app:new}
We can  easily show from \eqref{eq:Q} that each rigid transformation $\phi_j$ can be independently computed by minimizing the following criterion:
\begin{equation}
\sum_{ik}\Bigg[  \alphajik \int_{\yji} 
\beta_{jik}  (\yji )   \left(\|\phi_j(\yji)-\bmu_k\|^2_{\bSigma_k} \right) d\yji\Bigg], 
\label{eq:newbubu}
\end{equation}
with the notation $\sum_{ik} = \sum_{i=1}^{N_j}\sum_{k=1}^K$.
In \eqref{eq:newbubu}, the rigid transformation $\phi_j$ within the term $\|\phi_j(\yji)-\bmu_k\|^2_{\bSigma_k}$  is the transformation we aim to optimize. The terms $\beta_{ijk}$ and $\alpha_{jik}$ are computed based on an estimation of $\phi_j$ obtained in the previous iteration of the EM algorithm; therefore, they remain unoptimized.

To maintain simplicity in notation and to distinguish between the rigid transformation $\phi_j$ from the previous EM iteration and the one sought for optimization, an asterisk will be used as a superscript to denote optimized parameters, while no special notation will be applied to the other parameters.
\eqref{eq:newbubu} is then written as:

\begin{equation}
\sum_{ik}\Bigg[  \alphajik \int_{\yji} 
\beta_{jik}  (\yji )   \left(\|\phi_j^{\star}(\yji)-\bmu_k\|^2_{\bSigma_k} \right) d\yji\Bigg]
\label{eq:newanex}
\end{equation}

Expressing $\|\phi^{\star}_j(\yji)-\bmu_k\|^2_{\bSigma_k}$ as 
$\phi_j^{\star}(\yji)^T {\bSigma_k}^{-1} \phi_j(\yji) - 2 \bmu_k^T {\bSigma_k}^{-1} \phi^{\star}_j(\yji) + \bmu_k^T {\bSigma_k}^{-1} \bmu_k$, we can achieve an estimation of $\phi_j^{\star}$ by minimizing:

\begin{equation}
\sum_{ik} \alphajik  \Bigg[  \int_{\yji} 
\beta_{jik}  (\yji )  \phi_j(\yji)^T {\bSigma_k}^{-1} \phi^{\star}_j(\yji)    d\yji 
- 2 \bmu_k^T {\bSigma_k}^{-1}  \int_{\yji} \beta_{jik}(\yji )  \phi^{\star}_j(\yji) d\yji \Bigg]
\label{eq:newanexsdf}
\end{equation}

The computation of \eqref{eq:newanexsdf} can be achieved by employing the change of variables $\vji = \phi^{\star}_j(\yji)$.
In this regard, since $\phi^{\star}_j(\yji) = \phi^{\star}_j \circ \phi^{-1}_j \circ \phi_j (\yji)$, it is
noteworthy that Equation \ref{eq:beta} can also be expressed as:

	\begin{equation}
\label{eq:beta2}
\beta_{jik} (\yji) = \mN(\phi^{\star}_j(\yji);\phi^{\star}_j \circ \phi^{-1}_j(\hat{\y}_{jik}), {\R^{\star}_j} {\R^T_j} (\I-\W_{jik}) \bSigma_k {\R_j} {\R^{\star}_j}^T), 
\end{equation}
where $\R^{\star}_j$, and $\R_j$   respectively denote the rotation matrices associated with $\phi^{\star}_j$ and 
$\phi_j$. 

To simplify the notation, we define ${\bf S}_{jik}$ as the covariance matrix ${\R^T_j} (\I-\W_{jik}) \bSigma_k {\R_j}$, and as $\hat{\mathbf{y}}_{jik}^{inv}$ the point $\phi^{-1}_j(\hat{\y}_{jik})$. We obtain:
\begin{equation}
\label{eq:beta3}
\beta_{jik} (\yji) = \mN(\phi^{\star}_j(\yji);\phi^{\star}_j(\hat{\y}^{inv}_{jik}), {\R^{\star}_j} {\bf S}_{jik} {\R^{\star}_j}^T)
\end{equation}

By applying the change of variables $\vji = \phi^{\star}_j(\yji)$  to \eqref{eq:newanexsdf}, we can then use \eqref{eq:beta3} to obtain:
\begin{equation}
\begin{array}{l}
-\sum_{ik}  \alphajik \Bigg[  \displaystyle \int_{\vji} \mN(\vji);\phi^{\star}_j(\hat{\y}^{inv}_{jik}), {\R^{\star}_j} {\bf S}_{jik} {\R^{\star}_j}^T)
  \vji^T {\bSigma_k}^{-1} \vji    d\vji + \\
- 2 \bmu_k^T {\bSigma_k}^{-1}  \displaystyle \int_{\vji} \mN(\vji);\phi^{\star}_j(\hat{\y}^{inv}_{jik}), {\R^{\star}_j} {\bf S}_{jik} {\R^{\star}_j}^T) \vji d\vji \Bigg]
\end{array}
\label{eq:newanex2}
\end{equation}

If ${\bf x} \sim \mathcal{N}(\bmu;\bSigma)$, then the expectation of  $\bf x$, that is $\displaystyle \int_{\bf x} \mathcal{N}(\bmu;\bSigma){\bf x} d{\bf x}$ is $\bmu$. Moreover, the expectation of ${\bf x}^T \bf A x $ is $Tr({\bf A \Sigma}) + {\bmu}^T {\bmu}$. 
Consequently, \eqref{eq:newanex2} writes:
\begin{equation}
\sum_{ik}  \alphajik  \Bigg[  Tr({\bSigma_k}^{-1} {\R^{\star}_j} {\bf S}_{jik} {\R^{\star}_j}^T)) + [\phi^{\star}_j(\hat{\y}^{inv}_{jik})]^T
[\phi^{\star}_j(\hat{\y}^{inv}_{jik})] 
- 2 \bmu_k^T {\bSigma_k}^{-1}  \phi^{\star}_j(\hat{\y}^{inv}_{jik})  \Bigg]
\label{eq:final}
\end{equation}

Under the hypothesis that $\bSigma_k = \sigma_k^2 \I$, then
\begin{equation}
\bSigma_{k}^{-1} {\R^{\star}_j} {\bf S}_{jik} {\R^{\star}_j}^T = {\R^{\star}_j}  {\R^T_j} (\I-\W_{jik})  {\R^{\star}_j}^T. 
\end{equation}

Consequently, we have: 
\begin{equation}
\begin{array}{lll}
Tr(\bSigma_{k}^{-1} {\R^{\star}_j} {\bf S}_{jik} {\R^{\star}_j}^T) &=& Tr( {\R^{\star}_j}  {\R^T_j} (\I-\W_{jik})  {\R^{\star}_j}^T) \\
&=& Tr({\R^T_j} (\I-\W_{jik})). 
\end{array}
\end{equation}

Hence, the trace does not depend on $\phi^{\star}_j$, and $\phi^{\star}_j$ can be estimated by minimizing the following criterion: 
\begin{equation}
\sum_{ik} \alphajik \| \phi^{\star}_j(\hat{\y}^{inv}_{jik})-\bmu_k\|^2_{\bSigma_k}.
\end{equation}

\bibliographystyle{plain}
\bibliography{refs}
\end{document}